\documentclass{article}

\usepackage[main, final]{neurips_2026}

\usepackage[utf8]{inputenc} 
\usepackage[T1]{fontenc}    
\usepackage{hyperref}       
\usepackage{url}            
\usepackage{booktabs}       
\usepackage{amsfonts}       
\usepackage{nicefrac}       
\usepackage{microtype}      
\usepackage{xcolor}         

\usepackage{amsmath}
\usepackage{amssymb}
\usepackage{mathtools}
\usepackage{amsthm}

\usepackage{bbm}
\usepackage{multirow}
\usepackage{wrapfig}
\usepackage{bold-extra}

\usepackage{graphicx}
\usepackage{subcaption}
\usepackage{algorithm}
\usepackage{algpseudocode}

\newcommand{\algoname}{\text{LLE-MORL}}
\newtheorem{definition}{Definition}[section]
\newtheorem{theorem}{Theorem}[section]
\newtheorem{corollary}{Corollary}[section]
\newtheorem{proposition}{Proposition}[section]

\newtheorem{condition}{Condition}[section]
\newtheorem{assumption}{Assumption}[section]
\newtheorem{lemma}{Lemma}

\title{Local Policy Manifolds for Efficient \\ Multi-Objective Reinforcement Learning}

\author{%
  \setcounter{footnote}{1}
  Qiyue Xia \thanks{Institute of Perception, Action and Behaviour}\\
  School of Informatics \\
  University of Edinburgh\\
  \texttt{qiyue.xia@ed.ac.uk} \\
  \And
  Tianwei Wang \thanks{Artificial Intelligence and its Applications Institute}\\
  School of Informatics \\
  University of Edinburgh\\
  \texttt{tianwei.wang@ed.ac.uk} \\
  \And
  J.~Michael Herrmann \textsuperscript{$\dagger$}\thanks{Corresponding author}\\
  School of Informatics \\
  University of Edinburgh\\
  \texttt{michael.herrmann@ed.ac.uk} \\
}

\begin{document}

\maketitle

\begin{abstract}
Multi-objective reinforcement learning (MORL) aims at optimising several, often conflicting goals to improve the flexibility and reliability of RL in practical tasks. 
This is typically achieved by finding a set of diverse, non-dominated policies that form a Pareto front in the performance space. 
However, constructing such a policy set remains computationally demanding, as it requires finding not a single optimal policy, but a set of policies to cover a wide range of trade-offs among the objectives.
We introduce LLE-MORL, an approach that traces policy manifolds by utilising the local relationship between the high-dimensional policy parameter space and the performance space. 
This structured representation enables an efficient search within contiguous solution domains via locally linear extrapolation, allowing for the rapid generation of high-quality solutions without extensive retraining.
We also provide a theoretical analysis of the policy manifold structure in order to specify the applicability of the method.
Experiments across diverse continuous control domains demonstrate that LLE-MORL consistently achieves higher Pareto front quality and efficiency than state-of-the-art approaches.
\end{abstract}


\section{Introduction}
\label{sec1:intro}

Reinforcement Learning (RL) has shown great promise in complex decision-making problems, enabling significant advancements in a wide range~\citep{silver2016mastering, levine2016end}. 
In real-world scenarios, however, problems often feature multiple, often conflicting, objectives. 
Under this circumstance, multi-objective approaches provide flexibility in practical applications of reinforcement learning by providing a modifiable policy that can be adjusted according to changes of preference among a set of objectives~\citep{roijers2013survey, hayes2022practical}. This has fostered the development of the field known as multi-objective reinforcement learning (MORL).
Ideally, the modifiable policies developed within MORL allow for efficient adaptation, ensuring that a policy optimal for one set of preferences can be readily transformed to be optimal for a new set when those preferences change.
To prepare such a modifiable policy for application, three problems have to be solved: (i) The \emph{learning} problem involves the solution of an RL problem for each combination of preference parameters or at least for a representative subset of preferences. (ii) The \emph{representation} problem requires a parametrisation of the policies,
which typically results in either a discrete set of individual policies (common in population-based methods) or a single, continuously adaptable policy (prevalent in deep reinforcement learning approaches).
(iii) The \emph{selection} problem is to identify a suitable policy in the application which includes dynamic adjustments to preference drifts and possibly the decision whether a different policy should be invoked or whether further training is required to respond to a temporary detection of suboptimality. 



We consider these problems as a coherent task in order to reduce the computational burden of the learning problem and improve the structure of the policy representation. We hypothesise that if a continuous representation of policies can be found where similar preferences correspond to similar policy parameters, then small performance differences might be compensable with brief, targeted retraining. 
It is also anticipated that such a structured policy representation would benefit the selection problem, though this aspect is not the primary focus of our current study.

While a globally continuous mapping is an ideal, we notice that in non-trivial problems, the relationship between the performance space and parameter space of policies is not a simple, single continuous mapping but can be described by a family of locally continuous components~\citep{xu2020prediction, li2024find}. Our findings suggest that effectively exploring just a few of these components can be sufficient to achieve competitive performance in typical benchmark problems.  
To explore these components, we take a different perspective. Rather than searching for Pareto-optimal policies independently, we aim to trace the Pareto front manifold directly by exploring local structure in the high-dimensional policy parameter space. We formalise the Parameter-Performance Relationship (PPR), which captures the local correspondence between policy parameters and objective performance. 
This viewpoint enables an efficient strategy for Pareto front (see Sect.~\ref{secct_pareto}) approximation: starting from a small set of policies, we can expand along directions on the local policy manifold that preserve PPR and generate new candidate solutions via a training-free extension stage, establishing an explicit relationship between the policy parameter space and the reward space.

Based on this perspective, we develop a novel algorithm \algoname{} (Locally Linear Extrapolation for Multi-Objective Reinforcement Learning). 
Our experiments demonstrate that the proposed algorithm can achieve high-quality Pareto front approximations with notable sample efficiency compared to existing approaches.  
By focusing on local structure in the policy parameter space, \algoname{} reduces the need for exhaustive training while maintaining strong coverage and solution quality. 
Beyond efficiency, the structured use of the PPR reveals local policy manifold structure, supporting a clearer characterisation of trade-offs along the Pareto front.
Our code is publicly available at \url{https://github.com/AQQQQY/LLE_MORL}.





\section{Background}
\label{sec2:background}

\subsection{Multi-Objective Reinforcement Learning}
Multi-Objective Reinforcement Learning (MORL) extends the traditional RL framework to scenarios where agents must consider multiple, often conflicting objectives. This extension allows for more sophisticated decision-making models that mirror real-world complexities where trade-offs between competing goals, such as cost versus quality or speed versus safety, are common. 
To ground this notion formally, we represent a MORL problem as a
Multi-Objective Markov Decision Process (MOMDP)
which generalises the standard MDP framework to accommodate multiple reward functions, each corresponding to a different objective. 

\begin{definition} {Multi-Objective Markov Decision Process (MOMDP).}
A MOMDP is defined by the tuple \( 
(\mathcal{S},\mathcal{A},P,R,\gamma)
\), where \(\mathcal{S}\) is the state space, \(\mathcal{A}\) is the action space, \(P(s'|s,a)\) is the state transition probability, $R:\mathcal{S} \times \mathcal{A} \rightarrow \mathbb{R}^d$ is a vector-valued reward function with \(d\) as the number of objectives, specifying the immediate reward for each of the objectives, \(\gamma\) is the discount factor.
\end{definition}

The crucial difference between MOMDPs and traditional single-objective MDPs is the reward structure. While single-objective MDPs use a scalar reward function, MOMDPs feature a vector-valued reward function that outputs $d$-dim vector rewards $\mathbf{r}=R(s,a)$, which delivers distinct numeric feedback for each objective.
We denote a policy by \(\pi\).
At each timestep \(t\), the agent in state \(s_t \in \mathcal{S}\)
selects an action \(a_t \sim \pi(\cdot \mid s_t)\), transitions to a new state \(s_{t+1}\) with probability \(P(s_{t+1} \mid s_t, a_t)\), and receives a reward vector
\(
  \mathbf{r}_{t}
  = \bigl[(R_1(s_t,a_t),\,R_2(s_t,a_t),\,\dots,\,R_d(s_t,a_t)\bigr]).
\)
We define the discounted return vector by
\(
  \mathbf{G}_t
  = \sum_{k=0}^{\infty} \gamma^k\,\mathbf{r}_{t+k},
\)
and the multi-objective action-value function of a policy \(\pi\) for a given state-action pair \((s,a)\) by
\(
  \mathbf{Q}^{\pi}(s,a)
    = \mathbb{E}_{\pi}\bigl[\mathbf{G}_t \mid s_t = s,\,a_t = a\bigr]
\).
The goal of MORL is to find a policy \(\pi\) such that the expected return of each objective can be optimised. In practice, we trade off objectives via a
scalarisation function $f_{\boldsymbol{\omega}}(\mathbf{r})$, which produces a scalar utility using preference vector $\boldsymbol{\omega}\in\Omega$, where \(\Omega\) is the preferences space. The scalarisation function $f_{\boldsymbol{\omega}}(\mathbf{r})$ is used for mapping the multi-objective reward vector \(\mathbf{r}(s,a)\) to a single scalar. In this paper, we consider the linear scalarisation function
\(
f_{\boldsymbol{\omega}}(\mathbf{r}(s,a))=\boldsymbol{\omega}^\mathbf{T}\mathbf{r}(s,a)
\), 
where the preference vector $\boldsymbol{\omega}$ lies in the probability simplex,
i.e., \(\sum_{i=1}^d \omega_i = 1\) and \(\omega_i \ge 0\), 
which is commonly used in MORL literature \citep{yang2019generalized,felten2024multi}.
When the preference dimension \(d = 1\), the MOMDP reduces to a standard single-objective MDP  under this linear scalarisation setting.

\subsection{Pareto Optimality\label{secct_pareto}}
In multi-objective optimisation, the concept of optimality differs from the single-objective case. Typically, no single policy simultaneously maximises all objectives due to inherent trade-offs. Without any additional information about the user’s preference, there can now be multiple possibly optimal solutions. In the following, we introduce several useful definitions for possibly optimal policies.

\begin{definition} {Pareto optimality.}
A policy \(\pi\) is said to \emph{dominate} another policy \(\pi'\) if and only if
\(
\forall i \in \{1,\dots,d\}, \quad V_i^\pi(s) \geq V_i^{\pi'}(s), \quad\text{and}\quad \exists j,\, V_j^\pi(s) > V_j^{\pi'}(s),
\)
where \(V_i^\pi(s) = \mathbb{E}_\pi[\sum_{t=0}^\infty \gamma^t R_i(s_t,a_t)\mid s_0=s]\) denotes the expected discounted return for objective \(i\) under policy \(\pi\).
A policy \(\pi^*\) is \emph{Pareto optimal} if and only if it is not dominated by another policy. 
The set of all Pareto optimal policies forms the \emph{Pareto set}:
\(
\Pi_P = \{\pi \mid \pi \text{ is Pareto optimal}\}.
\)
The corresponding set of expected returns incurred by policies in the Pareto set is termed \emph{Pareto front}:
\(
\mathcal{P} = \{V^\pi(s) \mid \pi \in \Pi_P\}.
\)
\end{definition}

Since obtaining the true Pareto set is intractable in complex problems, the practical aim of multi‐objective optimisation is to construct a finite set of policies that closely approximate the true Pareto front.  So that practitioners can select the policy based on their preferred trade-off among objectives.

\section{Methods}
\label{sec3:method}

\begin{wrapfigure}{R}{0.5\textwidth}
    \vspace{-5em}
    \centering
    \includegraphics[width=\linewidth]{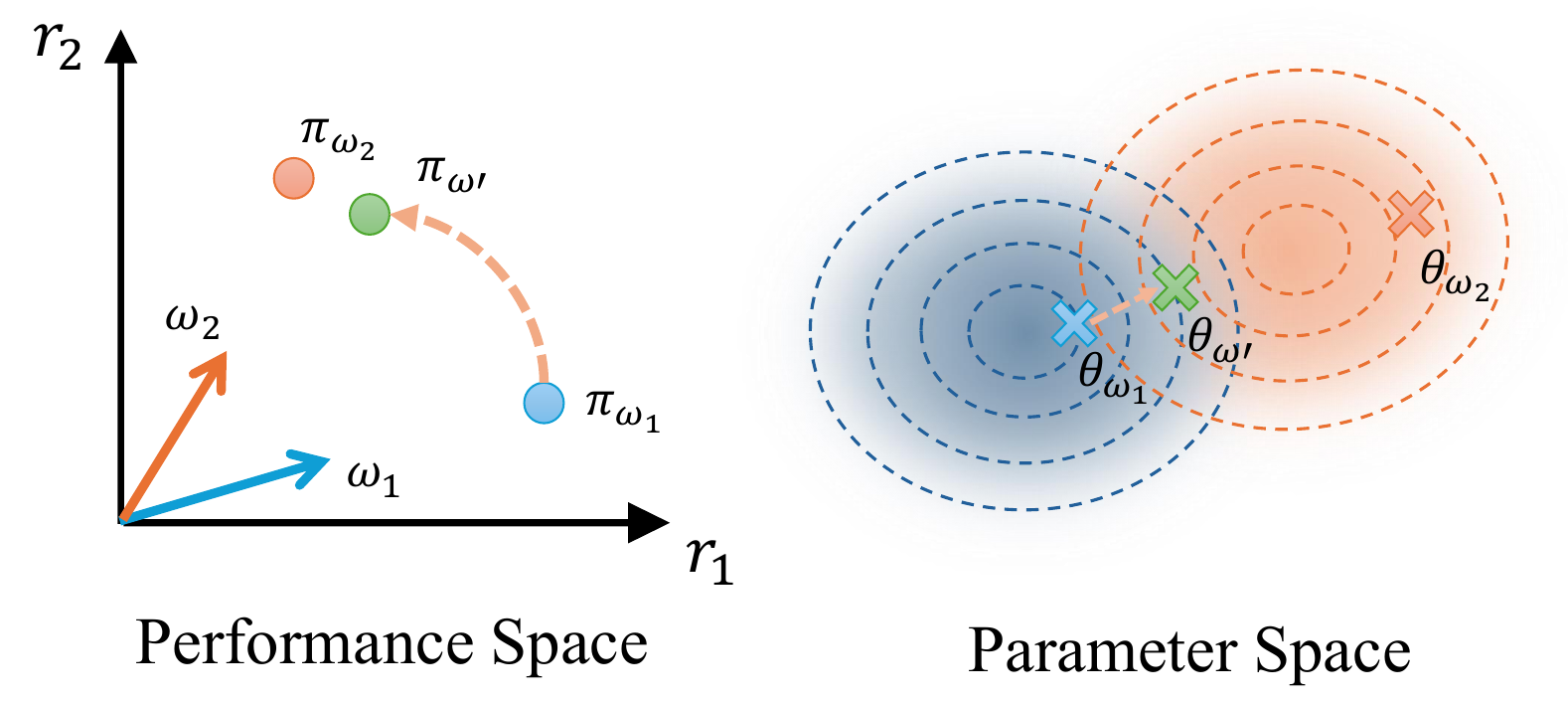}
    \caption{Retraining of a policy can produce policies that are similar in parameter space, but cover other regions in
    Pareto performance space.
    \label{fig:parameter_performance_space}}
\end{wrapfigure}

\subsection{Parameter-Performance Relationship\label{sec:PPR}}

Recent work in MORL has implicitly suggested a relationship between the parameter space of the policy network and the Pareto front in the performance space. 
~\citep{xu2020prediction} empirically show for PGMORL that each disjoint policy family occupies a continuous region in parameter space and maps to a contiguous segment of the Pareto front, while MORL/D 
~\citep{felten2024multi} assume that policies with similar parameters should lead to close evaluations. 
Motivated by these implicit observations, we formalise the \emph{parameter--performance relationship} and proceed to explain and empirically validate this property.

\begin{definition}{Parameter-Performance Relationship (PPR).}\label{PPR}
Let \(\Theta\subseteq\mathbb{R}^n\) be the space of policy parameters, and $V$ be the expected discounted return as a function $V:\Theta\to\mathbb{R}^d$ of the parameter $\theta$. We say that $V$ exhibits a \emph{parameter–performance relationship} on an open region \(U\subseteq\Theta\) if $\exists\delta>0$ and a function $h:U\times B_\delta(0)\to\mathbb{R}^d$, \emph{s.t.} $\forall \theta\in U$ and perturbation \(\Delta\theta\) satisfying $\|\Delta\theta\|<\delta$ and \(\theta+\Delta\theta\in U\), the difference in the performance space is:
\begin{equation}
  V(\theta+\Delta\theta)\;-\;V(\theta)
  \;=\;h\bigl(\theta,\Delta\theta\bigr).
\end{equation}
where $B_\delta(0)$ is an open ball with radius $\delta$ centred on the origin, i.e. $B_\delta(0):=\{x:\|x\|<\delta\}$.
\end{definition}

PPR is a local property and need not hold globally across the entire parameter space. In high-dimensional parameter spaces, policies with similar performance may be far apart in parameter space. This raises the question of how to identify policies that satisfy the PPR locally. As a first step, we require a metric for policy closeness in parameter space. 
We adopt the Hungarian matching distance~\citep{kuhn1955hungarian, munkres1957algorithms} to measure model distance and thereby quantify structural similarity between policies.
A formal definition of Hungarian matching distance is provided in Appendix~\ref{app:hungarian}. 
This metric naturally handles the permutation invariance of hidden units~\citep{goodfellow2016deep} and measures the smallest structural change needed to align one model to another: lower Hungarian distance indicates greater model similarity.


As illustrated in Figure \ref{fig:parameter_performance_space}, the red and blue gradient shadings and contour lines indicate the optimal region under different preference vectors \(\omega_1\) and \(\omega_2\) in high-dimensional parameter space and the arrow marks a short retraining update.
Although separately trained policies $\pi_{\omega_1}$ and $\pi_{\omega_2}$ under different preference weights $\omega_1$ and $\omega_2$ may be far apart in high-dimensional parameter space, a short retraining using $\omega_2$ starting from $\pi_{\omega_1}$ typically induces a small, structured parameter update together with a directional shift in the expected returns. 
This observation motivates our use of brief retraining as a practical means of obtaining locally related policies for which the PPR holds.

\begin{figure}[h]
\centering
    \begin{subfigure}[t!]{0.48\linewidth}
        \centering
        \includegraphics[width=\linewidth]{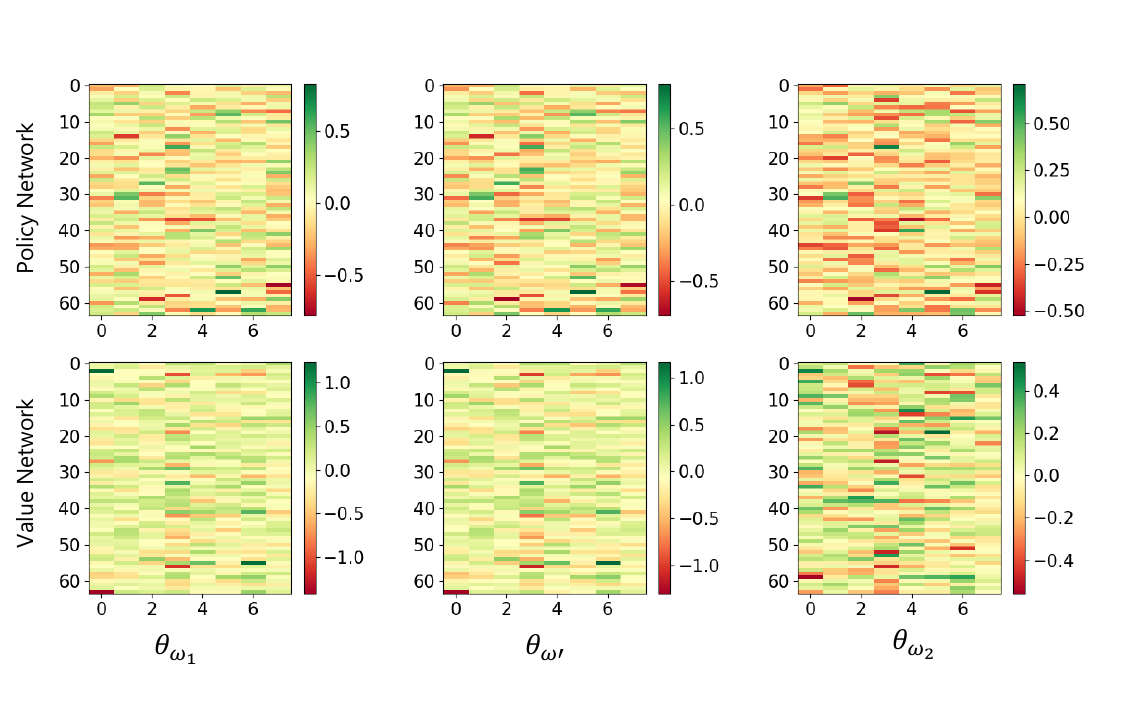}
        \caption{Heatmaps of policy‐net and value-net parameters to illustrate the retraining effect, compare Figure~\ref{fig:parameter_performance_space}.}
        \label{fig:net_diff}
    \end{subfigure}
    \hfill
    \begin{subfigure}[t!]{0.21\textwidth}
        \centering
        \includegraphics[width=\linewidth]{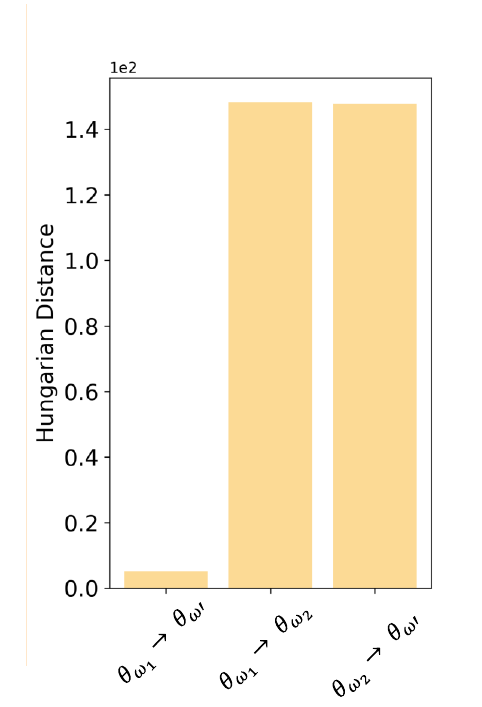}
        \caption{Combined Hungarian model distance.}
        \label{fig:param_diff_hungarian}
    \end{subfigure}
    \hfill
    \begin{subfigure}[t!]{0.26\textwidth}
        \centering
        \includegraphics[width=\linewidth]{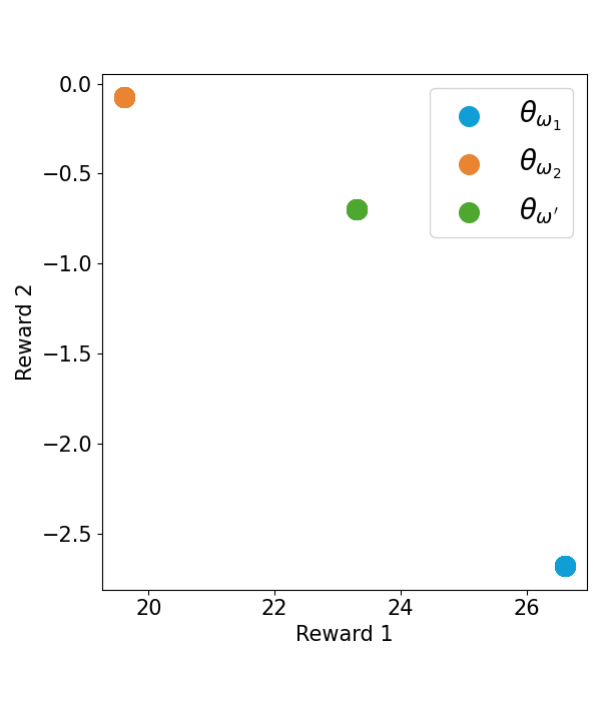}
        \caption{Respective positions in performance space.}
        \label{fig:reward_space_comparison}
    \end{subfigure}
    \caption{Visual comparison of fully trained policies $\theta_{\omega_1},\theta_{\omega_2}$ and briefly retrained policy $\theta_{\omega'}$.}
    \label{fig:model_similarity}
\end{figure}

\paragraph{Sanity Check\label{valhom}}
To obtain an initial empirical understanding of the PPR, we compare independently trained policies with those obtained via brief retraining.
We first train two policies to the stable stage using a multi-objective PPO-based~\citep{schulman2017proximal} algorithm under scalarization vectors \(\omega_1\) and \(\omega_2\), yielding model parameters \(\theta_{\omega_1}\) and \(\theta_{\omega_2}\).  Starting from \(\theta_{\omega_1}\), we perform a short retraining step under \(\omega_2\) to obtain \(\theta_{\omega'}\).  
We qualitatively evaluated the similarity between these policies using the policy and value network neuron heat maps in Figure \ref{fig:net_diff}, and quantitatively using the Hungarian matching distances between model pairs in Figure~\ref{fig:param_diff_hungarian}. In Figure \ref{fig:net_diff}, the horizontal axis corresponds to network layers, the vertical axis corresponds to neurons within each layer, and colour indicates the parameter values.
Their corresponding rewards are visualised in the two-objective performance space for the multi-objective {\sc Swimmer} problem in Figure~\ref{fig:reward_space_comparison}. 

We compare three pairs of models:  
(1)~\(\theta_{\omega_1}\) and \(\theta_{\omega_2}\), representing independently trained policies that differ substantially in both parameter and performance space;
(2)~\(\theta_{\omega_1}\) and \(\theta_{\omega'}\), showing that brief retraining produces a structurally similar policy with a low Hungarian distance while inducing a clear shift in performance; 
and (3)~\(\theta_{\omega'}\) and \(\theta_{\omega_2}\), demonstrating that although their parameters remain distinct, their rewards lie much closer in the performance space.  
These empirical observations indicate that a short retraining under a new preference produces a small, structured parameter update that directly maps to a predictable shift in performance, validating the PPR. 
More details of the sanity check procedure are provided in Appendix \ref{app:sanity_check}.

\subsection{Locally Linear Extension}
Based on the PPR formalisation, a natural question is whether the parameter-space difference between two policies that satisfy the PPR can serve as a directional update for tracing the Pareto front along a local policy manifold.
To explore this, we consider a base policy \(\theta_{\omega}\) and a retrained policy \(\theta_{\omega'}\) that exhibit a PPR. 
For this directional information to be meaningful for Pareto front exploration, both \(\theta_{\omega}\) and  \(\theta_{\omega'}\) should ideally be non-dominated solutions, at least with respect to each other.
Given such a pair, we compute the parameter update vector \(\Delta\theta = \theta_{\omega'} - \theta_{\omega}\) and generate intermediate policies by moving from the base policy \(\theta_\omega\) along the parameter displacement \(\Delta\theta\) in scaled steps. 
Concretely, for each scale \(\alpha\), we form 
\(\theta_\alpha = \theta_\omega + \alpha\,\Delta\theta\)\, and evaluate its reward vectors in preference space.

\begin{figure*}[ht]
    \centering
    \begin{subfigure}[t]{0.32\textwidth}
        \includegraphics[width=\linewidth]{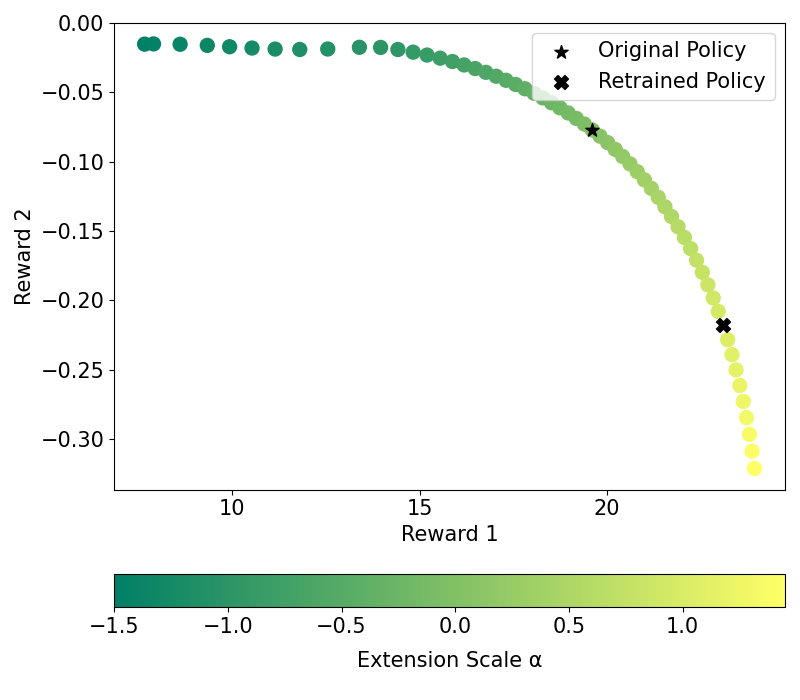}
        \caption{}
    \end{subfigure}\hfill
    \begin{subfigure}[t]{0.32\textwidth}
        \includegraphics[width=\linewidth]{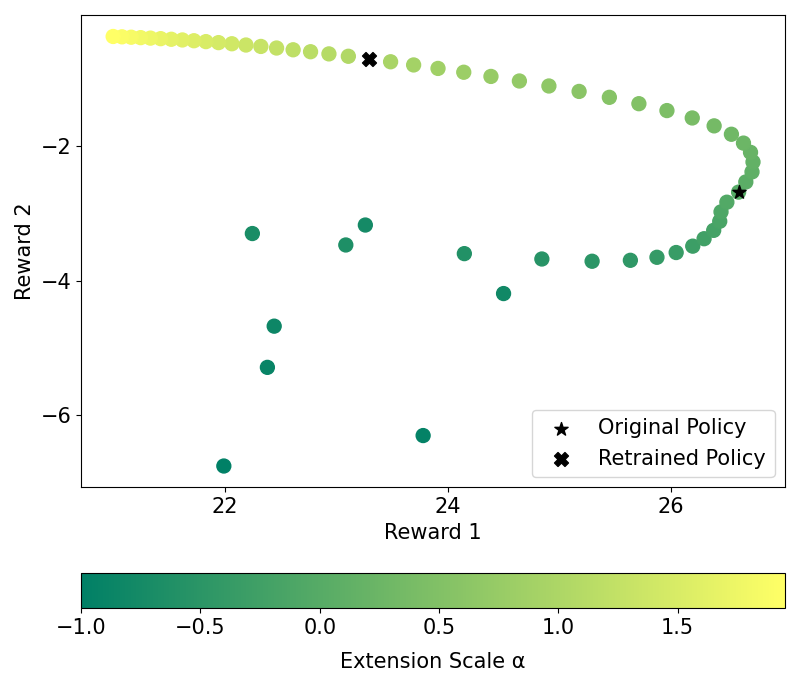}
        \caption{}
    \end{subfigure}\hfill
    \begin{subfigure}[t]{0.32\textwidth}
        \includegraphics[width=\linewidth]{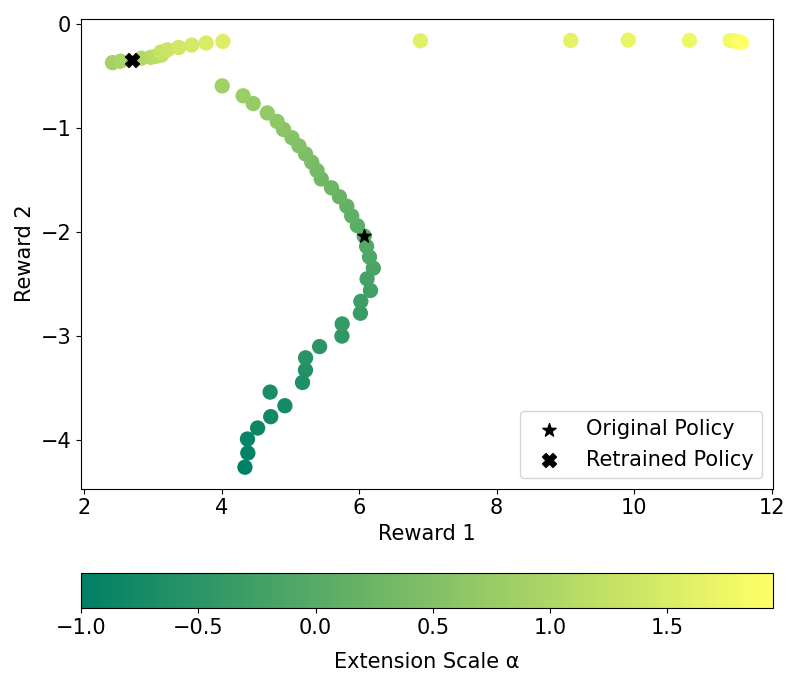}
        \caption{}\label{fig:moving_c}
    \end{subfigure}
    
    \caption{Visualisation of the process of applying the parameter difference
    \( \Delta\theta = \theta_{\omega'} - \theta_{\omega}\) between two related policies. The policies are obtained by first training a policy~\(\theta_{\omega}\) to a stable stage using a scalarization vector~\(\omega\) and then finding a policy~\(\theta_{\omega'}\) by a brief additional training period with a different scalarization vector~\(\omega'\). Iterating the shift $\Delta\theta$ in the policy space induces a sequence of shifts also in the multi-objective reward space.
    The subfigures show results for 
    different initial preferences: (a) A convex front is found from the two policies. 
    (b) Although the original policy turns out to be Pareto suboptimal, the solution manifold extends into a Pareto optimal component. (c) Retraining can cause the (Pareto-suboptimal) original solution to jump to a different branch so that the corresponding solution consists of two components one of which can be ignored because of Pareto suboptimality.\label{fig:moving}}
\end{figure*}

Figure \ref{fig:moving} visualises the resulting trajectory of reward vectors in the two-dimensional objective space: as \(\alpha\) grows, the trajectory passes through the region around \(\theta_{\omega'}\) and can extend beyond both the base and retrained endpoints, demonstrating how simple parameter‐space moves can 
traverse broad trade‑off regions, which offers a cost‑effective strategy for efficiently expanding an approximate Pareto front without training each point from scratch. 

\subsection{The \algoname{} Algorithm}\label{subsec:lle_algorithm}
Motivated by empirical observations, 
\algoname{} is built on the hypothesis that there exist local policy manifolds in parameter space that correspond locally to the Pareto front and have the same intrinsic dimension, along which multi-objective performance varies smoothly. Theoretical analysis and formal proofs are provided in Section \ref{subsec:theoretical} and Appendix \ref{app:theory}. 

Consider a $d$-objective MORL problem. The full Algorithm \ref{alg:pareto_extension} (see Appendix \ref{app:algorithm}) consists of five stages:
\textbf{(1)~Initialisation:} We train a set of \(K\) base policies \(\{\theta_{\omega_k}\}_{k=1}^K\) to stable using PPO~\citep{schulman2017proximal}. Each policy is trained under a distinct scalarization weight \(\omega_k\in\Omega\), where these weights are chosen to be evenly distributed across the preference space.
\textbf{(2)~Directional Retraining:} For each base policy $\theta_{\omega_k} (k=1,\dots,K)$, 
we estimate a set of $m=d-1$ local directions on the policy manifold in parameter space by retraining.
Specifically, for each $i=1,\dots,m$, we continue training from $\theta_{\omega_k}$ under a nearby preference $\omega_k^{(i)}$ for $T_{\mathrm{dir}}$ steps to obtain a directional policy $\theta_{\omega_k^{(i)}}$, where $\theta_{\omega_k}$ and $\theta_{\omega_k^{(i)}}$ are all non-dominated.
We record $ \Delta\theta_k^{(i)} = \theta_{\omega_k^{(i)}} - \theta_{\omega_k}$, 
and $\Delta \omega_k^{(i)} = \omega_k^{(i)} - \omega_k$ as the parameter update vector and the corresponding preference shift.
\textbf{(3)~Locally Linear Extension:} 
For each base policy $\theta_{\omega_k}$, we generate candidate policies by extrapolating along the $m$ local directions using $\{\Delta\theta_k^{(i)}\}_{i=1}^{m}$.
Each candidate is constructed as a linear combination of these directions:
$\theta_{k,\mathrm{cand}}=\theta_{\omega_k}+\sum_{i=1}^m\alpha_i\Delta\theta_k^{(i)}$, 
where step-scale factors $\alpha$ control the step sizes along each direction.
We choose each $\alpha_i$ from a discrete grid defined by $\alpha_{\mathrm{start}}$, $\alpha_{\mathrm{end}}$, and step size $\Delta\alpha$, resulting in $M$ possible values per direction.
Iterating over all combinations across the $m$ directions produces $M^{m}$ candidate policies for each base solution.
In parallel, we assign each candidate a matched preference weight $\omega_{k,\mathrm{cand}}=\omega_k + \sum_{i=1}^{m}\alpha_i\,\Delta \omega_k^{(i)}$.
\textbf{(4)~Candidate Selection:} All candidates $\theta_{k,\mathrm{cand}}$ generated in the locally linear extension stage are evaluated to obtain their respective performance vectors. 
We then select the subset of non-dominated candidates and forward them to the fine-tuning stage. 
\textbf{(5)~Preference-Aligned Fine-Tuning:}
For each selected candidate $\theta$ and its matched weight $\omega$, we perform a short PPO fine-tuning phase of $T_{\mathrm{ref}}$ steps under $\omega$ to push the policy closer to the true Pareto front.

Together, these stages explore local policy manifold structure in parameter space, enabling efficient tracing of the local Pareto front manifold. 
This construction yields a structured policy representation: local directions in parameter space correspond to meaningful objective trade-offs, allowing parameter changes to be directly associated with movements along the Pareto front.

\subsection{Theoretical Analysis}\label{subsec:theoretical}
In this section, we present a theoretical analysis for the effectiveness of \algoname{}, formally modelling the local manifold structure of Pareto-optimal policies and establishing guarantees for the locally linear extension mechanism used to trace the Pareto front.
For clarity, we only provide the theory framework here, while proofs are located in Appendix \ref{app:theory}. First, the continuity of the reconstructed Pareto front by \algoname{} is guaranteed by the Lipschitz continuity of PPR. Following Theorem \ref{theo:lc} and Corollary \ref{theo:cor:sep} provide the theoretical foundation of these properties. 
\begin{theorem}{(Lipschitz Continuity of PPR function $h$.)}\label{theo:lc}
Denote the expected discounted return by $V:U\rightarrow\mathbb{R}^d$, where $U\subseteq\Theta\subseteq\mathbb{R}^n$ is an open set. The sufficient conditions for Lipschitz continuity of $h(\theta,\Delta\theta)$ with respect to the second variable $\Delta\theta$ are: $V(\theta)$ is differentiable on $U$, and the score function is bounded, i.e. $\forall a\in\mathcal{A}, s\in\mathcal{S}$, there is 
\begin{equation}
\sup_{\theta\in U} \bigl\|\nabla_\theta \log \pi_\theta(a|s)\bigr\|\le \mathcal{G}<\infty.
\label{theo:scorefunction}
\end{equation}
\end{theorem}
The inequality Equation (\ref{theo:scorefunction}) states that the score function $\nabla_\theta \log \pi_\theta(a|s)$ introduced in \cite{baxter2001policy-gradient} is bounded by a constant $\mathcal{G}$. This is a common regularity assumption in related RL research, which can be found in \cite{aydin2023policy-gradient,shi2024policy-gradient}.

\begin{corollary}{(Surface continuity guaranteed by Lipschitz PPR)}\label{theo:cor:sep}
Algorithm LLE can construct a continuous surface in the performance space, as an approximation to the Pareto front, given appropriate regularity conditions and a range of $\alpha$.
\end{corollary}
Further, we aim to evaluate the error between the reconstructed result by \algoname{} and the Pareto front $\mathcal{P}$. Assume $\mathcal{P}$ is a differentiable manifold of dimension $m$ embedded in the Euclidean space of dimension $d$, i.e. $\mathcal{P}\in \mathcal{M}^m(\mathbb{R}^d)$, where $\mathrm{dim}\mathcal{P}$ is $m=d-1$.

Suppose the initialisation returns $\theta^{(0)}$, assuming it corresponds to a non-dominated performance, where $v:=V\big(\theta^{(0)}\big)\in\mathcal{P}$. Let $P_v\subset\mathcal{P}$ be an open neighbourhood of $v$ on the manifold $\mathcal{P}$, where $P_v\in\mathcal{M}^m(\mathbb{R}^d)$. Define the neighbour of $v$ to be $P_v$, assume there exists a parameter set $\Theta_v\subset\Theta$ bijectively mapping to $P_v$, denoted by $V|_{\Theta_v}:\Theta_v\rightarrow P_v$.
\begin{theorem}{(Local lifting manifold)}\label{theo:thm:smooth_lift}
  If $V$ is differentiable on $U$ and $(V|_{\Theta_v})^{-1}\!:\!P_v\!\!\to\mathbb{R}^n$ is $C^k$,
  then $\Theta_v$ is a differentiable manifold embedded in $\mathbb{R}^n$ of the same dimension as $P_v$, i.e. $\Theta_v\in\mathcal{M}^m(\mathbb{R}^n)$.
\end{theorem}
Therefore, the relation between the Pareto front and corresponding parameters can be understood as the diffeomorphism between two differentiable manifolds of the same dimension $m$. Next, we define the expression of linear extension. 
\begin{condition}{(Local extension directions)}\label{theo:con:directions_hd}
Suppose the directional retraining in \algoname{} returns $\theta^{(i)}\in\Theta_v$, $i=1,\dots,m$. Define
\begin{equation}
\mathcal{D}:=\bigl[\theta^{(1)}-\theta^{(0)},\ \dots,\ \theta^{(m)}-\theta^{(0)}\bigr]\in\mathbb{R}^{n\times m}
\end{equation}
which has full column rank, i.e. $\mathrm{rank}(\mathcal{D})=m$. Further, $\mathrm{span}(\mathcal{D})$ provides an approximation to the tangent space of $\Theta_v$ at $\theta^{(0)}$, i.e. $T_{\theta^{(0)}}\Theta_v$.
\end{condition}

\begin{theorem}{(Local Pareto error of LLE)}\label{theo:theorem:lle} Denote the local linear extrapolation by $\tilde\theta(\alpha):=\theta^{(0)}+\mathcal{D}\alpha$. Then, for a suitable range of $\alpha\in\mathbb{R}^m$, the error of extrapolation to the Pareto front is bounded, with first-order convergence as $\|\alpha\|\rightarrow0$ and second-order increase as $\|\alpha\|$ grows. The error of exploration to the actual Pareto front is controlled by:
\begin{equation}
\mathrm{dist}\bigl(V(\tilde\theta(\alpha)),\,\mathcal{P}_v\bigr)
\le \frac{L_V \eta}{2}\max_{1\leq i\leq m}\|x^{(i)}\|^2\Big(\sqrt m\,\|\alpha\|+m\,\|\alpha\|^2\Big).
\label{theo:eq:error_bound}
\end{equation} 
\end{theorem}

The detailed proof of this error bound can be found in Theorem \ref{theorem:lle} in Appendix \ref{app:theory}.
Theorem \ref{theo:theorem:lle} supports the effectiveness of \algoname{} in constructing the Pareto front. And the error is controlled by an appropriate range of $\alpha$, corresponding to the distance of linear extension. The empirical results and discussion are stated in Appendix \ref{app:exp_alpha} for the feasibility of the choice of $\alpha$. 

\section{Experiments}
\label{sec4:experiment}

\begin{figure}[!t]
  \centering 
  \captionsetup[subfigure]{font=footnotesize}
  \begin{subfigure}[t]{0.32\linewidth}
    \centering
    \includegraphics[width=\linewidth]{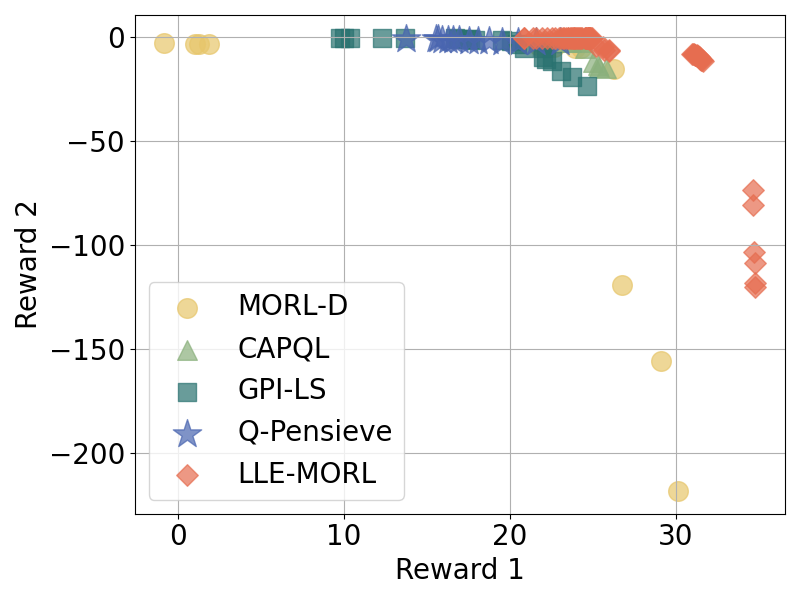}
    \caption{\centering MO-Swimmer (sample-efficient)}
    \label{fig:pf_swimmer}
  \end{subfigure}
  \hfill
  \begin{subfigure}[t]{0.32\linewidth}
    \centering
    \includegraphics[width=\linewidth]{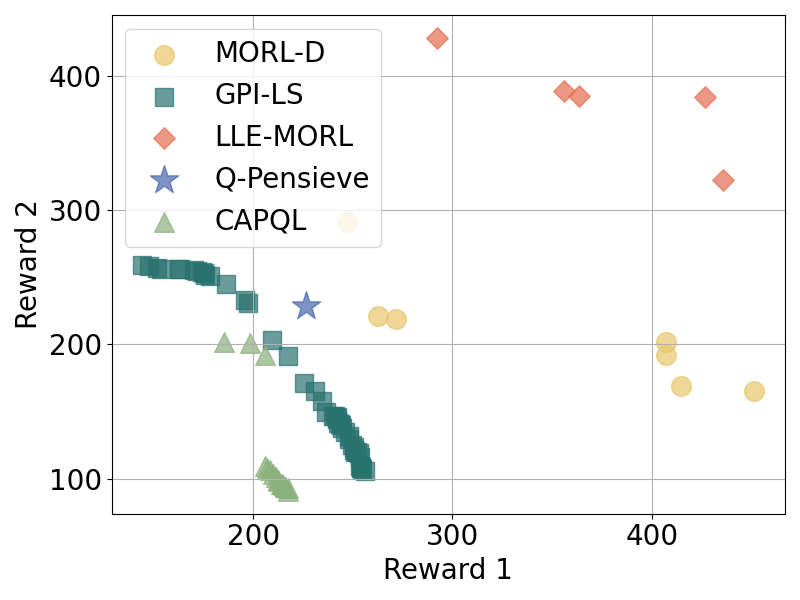}
    \caption{\centering MO-Hopper-2d (sample-efficient)}
    \label{fig:pf_hopper}
  \end{subfigure}
  \begin{subfigure}[t]{0.32\linewidth}
    \centering
    \includegraphics[width=\linewidth]{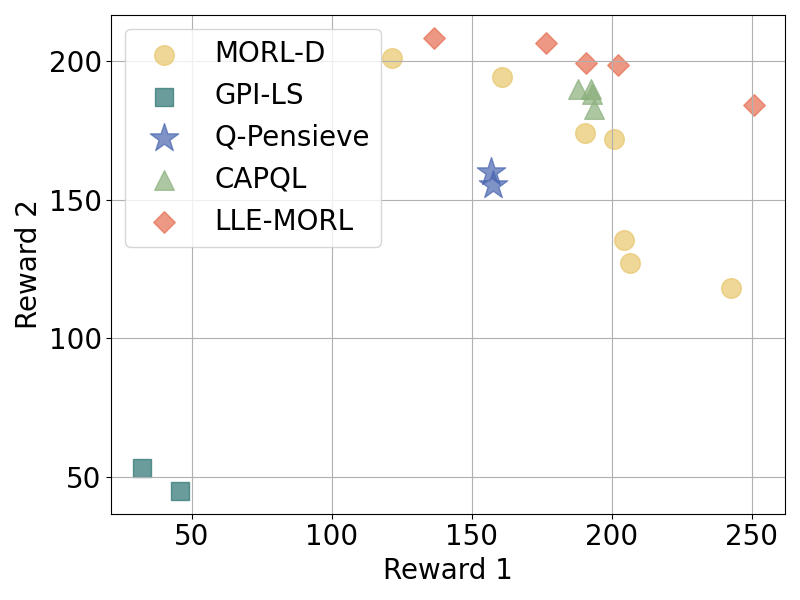}
    \caption{\centering MO-Ant-2d (sample-efficient)}
    \label{fig:pf_ant}
  \end{subfigure}

  \hfill

  \begin{subfigure}[t]{0.32\linewidth}
        \centering
        \includegraphics[width=\linewidth]{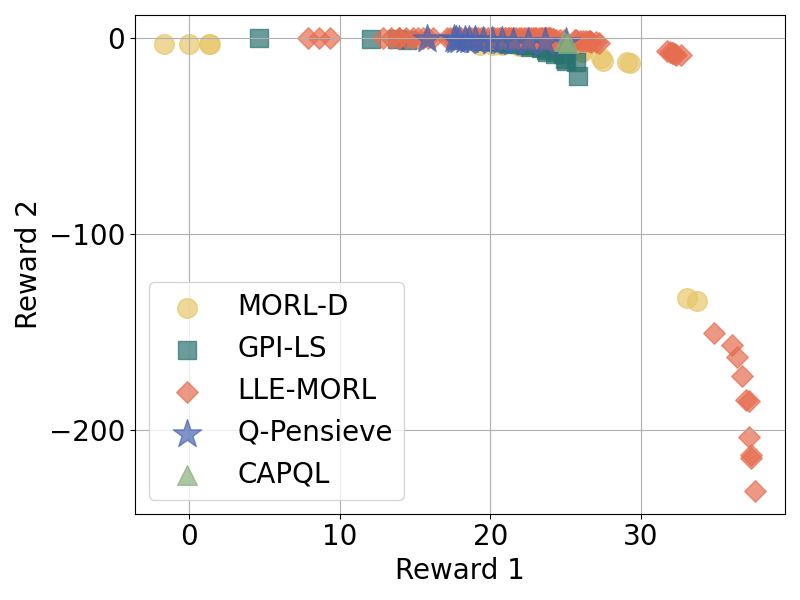}
        \caption{\centering MO-Swimmer (standard-training)}
        \label{fig:pf_swimmer_full}
  \end{subfigure}
  \hfill
  \begin{subfigure}[t]{0.32\linewidth}
        \centering
        \includegraphics[width=\linewidth]{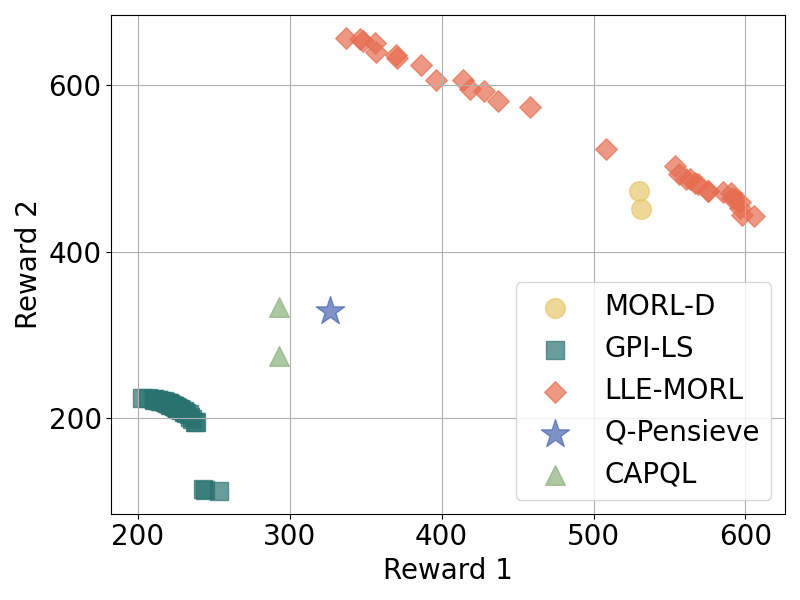}
        \caption{\centering MO-Hopper-2d (standard-training)}
        \label{fig:pf_hopper_full}
      \end{subfigure}
    \hfill
    \begin{subfigure}[t]{0.32\linewidth}
        \centering
        \includegraphics[width=\linewidth]{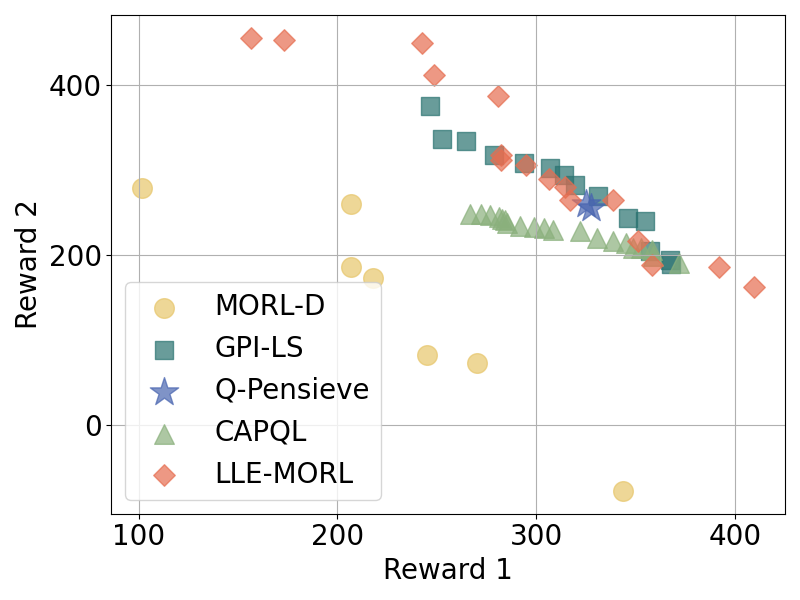}
        \caption{\centering MO-Ant-2d (standard-training)}
        \label{fig:pf_ant_full}
    \end{subfigure}

    \caption{
    Pareto fronts from 2 different settings, comparing our \algoname{} method with baselines on three 2d continuous-control benchmarks.
    \algoname{} consistently achieves wider coverage and closer proximity to the true Pareto front. }
    \label{fig:pareto_fronts_2d}
\end{figure}

\subsection{Experiment Setup}
In this section, we evaluate the \algoname{} algorithm using popular continuous MORL benchmark problems from the MO-Gymnasium-v5 \citep{felten_toolkit_2023}. 
Our benchmark problems include three two-objective continuous environments: \textbf{MO-Swimmer}, \textbf{MO-Hopper-2d}, \textbf{MO-Ant-2d}, and two three-objective continuous environments: \textbf{MO-Hopper-3d}, \textbf{MO-Ant-3d}.
We evaluate the quality of the approximate Pareto front using three standard metrics: \textbf{Hypervolume (HV)}, \textbf{Expected Utility (EU)}, \textbf{Sparsity (SP)}, following the formalism in~\citep{zitzler2002multiobjective, zintgraf2015quality, hayes2022practical}.
Higher HV and EU values indicate better overall front quality, while SP measures the distribution of solutions along the front. Since SP is scale-dependent and less directly related to decision quality, we report it mainly as a complementary metric to assess coverage diversity. More details can be found in Appendix \ref{app:exp_details}.

We compare our \algoname{} against the following state-of-the-art MORL algorithms:  
(i) \textbf{GPI-LS} \citep{alegre2023sample}
applies Generalised Policy Improvement over a discretised set of preference weights and uses linear scalarization to construct a diverse Pareto set.
(ii) Concave-augmented Pareto Q-learning (\textbf{CAPQL}) \citep{lu2023multi}:
 learns an ensemble of Q-functions under different preferences and selects actions via conservative aggregation to improve front coverage.
(iii) \textbf{Q-Pensieve} \citep{hung2023qpensieve}  boosts the sample efficiency of MORL by storing past Q-function snapshots in a replay buffer, enabling explicit policy-level knowledge sharing across training iterations.
(iv) \textbf{MORL/D} \citep{felten2024multi} is a deep-RL analogue of decomposition-based multi-objective optimisation that trains subpolicies under scalarised objectives and recombines them via weight decompositions to approximate the Pareto front. 
We additionally report \algoname{}-0, which disables the fine-tuning stage of LLE-MORL, as an ablation variant; the corresponding analysis is provided in Section~\ref{subsec:ablation}.

\subsection{Results and Analysis on 2D Benchmarks}

\begin{table*}
    \centering
    \caption{Evaluation of the Pareto front quality on 2D-objective benchmarks under different settings.}
      \scalebox{0.65}{
  \begin{tabular}{c| l| l| cccc cc}
    \toprule
    \multirow{2}{*}{Setting}
      & \multirow{2}{*}{Environment} 
      & \multirow{2}{*}{Metric} 
      & \multicolumn{6}{c}{Method} \\
    \cmidrule(lr){4-9}
     &  &  & GPI-LS & CAPQL & Q-Pensieve & MORL/D & \algoname-0 &\algoname \\
    \midrule
    \multirow{9}{*}{Sample-efficient} 
    & \multirow{3}{*}{MO-Swimmer} 
      & HV\((10^4)\) & 4.80$\pm$0.43 & 5.00$\pm$0.09 & 4.77$\pm$0.17 & 5.72$\pm$0.25 & 6.54$\pm$0.10 & \textbf{6.77$\pm$0.17} \\
      & & EU\((10^{1})\) & 0.89$\pm$0.14 & 1.08$\pm$0.01 & 1.06$\pm$0.11 & 1.00$\pm$0.01 & 0.93$\pm$0.11 & \textbf{1.11$\pm$0.02}\\
      & & SP\((10^{2})\) & 1.12$\pm$1.02 & 1.68$\pm$1.33 & \textbf{0.82$\pm$0.61} & 11.03$\pm$7.17 & 1.61$\pm$1.21 & 1.78$\pm$1.60\\
    \cmidrule(lr){2-9}
    & \multirow{3}{*}{MO-Hopper-2d} 
      & HV\((10^5)\) & 1.06$\pm$0.12 & 1.40$\pm$0.31 & 1.00$\pm$0.05 & 1.96$\pm$0.19 & 2.73$\pm$0.17 & \textbf{3.02$\pm$0.28}\\
      & & EU\((10^{2})\) & 2.11$\pm$0.17 & 2.68$\pm$0.40 & 2.16$\pm$0.08 & 3.29$\pm$0.15 & 4.15$\pm$0.20 & \textbf{4.36$\pm$0.28}\\
      & & SP\((10^{2})\) & 5.35$\pm$3.64 & 1.33$\pm$1.21 & \textbf{0.08$\pm$0.04} & 48.25$\pm$22.90 & 8.08$\pm$6.44 & 15.13$\pm$8.31\\
    \cmidrule(lr){2-9}
    & \multirow{3}{*}{MO-Ant-2d} 
      & HV\((10^4)\)  & 3.24$\pm$1.28 & 8.10$\pm$0.64 & 5.01$\pm$1.73 & 9.73$\pm$0.42 & 10.07$\pm$0.63 & \textbf{10.44$\pm$0.56}\\
      & & EU\((10^{2})\)  & 0.72$\pm$0.30 & 1.82$\pm$0.10 & 1.18$\pm$0.41 & 2.03$\pm$0.08 & 2.16$\pm$0.10 & \textbf{2.23$\pm$0.08}\\
      & & SP\((10^{3})\)  & 1.14$\pm$1.01 & \textbf{0.05$\pm$0.03} & 0.12$\pm$0.10 & 1.33$\pm$1.02 & 1.51$\pm$0.36 & 1.22$\pm$0.22\\
    \midrule

    \multirow{9}{*}{Standard-training} 
    & \multirow{3}{*}{MO-Swimmer} 
      & HV\((10^4)\) & 5.46$\pm$0.17 & 5.12$\pm$0.32 & 5.20$\pm$1.08 &7.05$\pm$0.39 & 7.74$\pm$0.22& \textbf{7.82$\pm$0.24} \\
      & & EU\((10^{1})\) &1.06$\pm$0.03 &1.10$\pm$0.05& 1.13$\pm$0.15 &1.13$\pm$0.03 &1.03$\pm$0.07 & \textbf{1.16$\pm$0.09} \\
      & & SP\((10^{2})\) & \textbf{0.04$\pm$0.02}& 0.12$\pm$0.09& 0.05$\pm$0.02 &4.36$\pm$1.64& 1.11$\pm$0.56& 1.15$\pm$0.63 \\
    \cmidrule(lr){2-9}
    & \multirow{3}{*}{MO-Hopper-2d} 
      & HV\((10^5)\) &1.15$\pm$0.04 & 2.11$\pm$0.91& 1.62$\pm$0.20 &3.75$\pm$0.25 & 4.76$\pm$0.08& \textbf{4.87$\pm$0.09} \\
      & & EU\((10^{2})\) & 2.26$\pm$0.07&3.46$\pm$1.07& 2.76$\pm$0.84 & 4.98$\pm$0.17& 5.67$\pm$0.10& \textbf{5.74$\pm$0.07} \\
      & & SP\((10^{2})\) &0.95$\pm$0.73 & 15.37$\pm$10.67& \textbf{0.02$\pm$0.01} &10.60$\pm$6.31 &5.09$\pm$1.77 & 4.53$\pm$1.79 \\
    \cmidrule(lr){2-9}
    & \multirow{3}{*}{MO-Ant-2d} 
      & HV\((10^5)\) &1.72$\pm$0.93 &1.63$\pm$0.72 & 1.78$\pm$0.34 &1.84$\pm$0.31 & 2.58$\pm$0.22& \textbf{2.68$\pm$0.22}\\
      & & EU\((10^{2})\) &2.68$\pm$1.19 &2.81$\pm$0.80 & 2.18$\pm$0.41 &3.09$\pm$0.30 & 3.81$\pm$0.27& \textbf{3.93$\pm$0.26} \\
      & & SP\((10^{3})\) & 0.57$\pm$0.50& 0.28$\pm$0.24& \textbf{0.10$\pm$0.09} &3.92$\pm$2.81 &2.23$\pm$0.97 & 1.59$\pm$0.40 \\
    \bottomrule

  \end{tabular}
  }

    \label{tab:metrics_2d}
\end{table*}

We now present quantitative results on 2D-objective benchmarks.
We conduct experiments under two distinct settings to provide a comprehensive understanding of these algorithms' capabilities: 
\textbf{(1)~Sample-Efficient Setting:} All methods, including our \algoname{}, were trained under an interaction budget for \(1.5\times10^5\) timesteps (including all algorithmic training phases). Given the complexity of continuous control benchmarks, this relatively limited interaction budget serves as a critical testbed for evaluating how rapidly different MORL strategies can discover effective Pareto front approximations. \textbf{(2)~Standard-Training Setting:} To assess performance under more common training conditions, all methods, including our \algoname{}, were trained under an interaction budget for \(1\times10^6\) timesteps. 
This setting aligns with common practices for benchmarking in continuous control and allows us to assess the final quality of the Pareto fronts achieved by each algorithm after a more thorough learning process.
In all experiments, we fix hyperparameters according to a common setting shared across methods, without assuming any task-specific parameter tuning. Full implementation details are provided in Appendix \ref{app:exp_details}.

In the sample-efficient setting, shown in Figure~\ref{fig:pareto_fronts_2d} and Table~\ref{tab:metrics_2d} top part, \algoname{} achieves the highest HV and EU in all benchmarks, 
demonstrating strong capabilities in rapidly achieving high-quality Pareto fronts.
Regarding SP, while \algoname{} does not consistently achieve the leading scores on this metric, its performance generally reflects a good and effective distribution of solutions along the high-quality Pareto fronts it identifies. 
It should be noted that SP results can be confounded by a fragmentary recovery of the Pareto front. 
For instance, if Q-Pensieve discovers only two close points of the Pareto front for the MO-Ant problem, the sparsity rating is nearly perfect. Meanwhile, a low SP might also arise from solutions being overly clustered in a small region, as potentially seen with GPI-LS and Q-Pensieve in MO-Swimmer shown in Figure \ref{fig:pf_swimmer}.

In the standard-training setting, shown in Figure~\ref{fig:pareto_fronts_2d} and Table~\ref{tab:metrics_2d} bottom part, across all benchmarks, \algoname{} achieves the highest HV and highly competitive EU, which suggests a more extensive and higher-quality approximation of the Pareto front than the baselines.
Pareto front visualisation in Figure~\ref{fig:pf_swimmer_full} further supports this result: on MO-Swimmer, LLE-MORL identifies solutions in the lower-right region that are missed by baselines such as GPI-LS, Q-Pensieve, and CAPQL.
Notably, when comparing the GPI-LS, CAPQL and Q-Pensieve performance to those in the sample-efficient setting for the MO-Swimmer environment, these particular baselines appear to remain constrained by suboptimal solutions in this challenging region, indicating that simply extending training duration did not resolve their exploration deficiencies here. 
While \algoname's thorough exploration to achieve this broader coverage means its SP may not be the numerically lowest, the result could be well-justified by the extensive nature of the front.

Overall, \algoname{} consistently achieves high-quality Pareto fronts across different settings. 
Its innovative training-free extension step traces the local policy manifold structure to efficiently generate diverse solutions under limited budgets and consistently improved results with extended training.
Runtime comparisons are reported in Appendix~\ref{app:exp_details}.

\subsection{Performance and Scalability in Higher Dimensional Objective Spaces}
To evaluate the robustness and discuss the scalability of LLE-MORL beyond standard 2D-objective trade-offs, we extended our experimental evaluation to higher-dimensional scenarios using the MO-Hopper-3d and MO-Ant-3d benchmarks. In these environments, the Pareto front is no longer a one-dimensional curve but a two-dimensional manifold embedded in a three-dimensional performance space. All methods, including our \algoname{} approach, were trained under an interaction budget for \(2\times10^6\) timesteps (including all algorithmic training phases).

\begin{table*}[t]
    \centering
    \caption{Evaluation of the quality of the Pareto front on 3D-objective benchmarks.}
    
  \scalebox{0.65}{
  \begin{tabular}{l| l| cccc cc}
    \toprule
    \multirow{2}{*}{Environment} 
      & \multirow{2}{*}{Metric} 
      & \multicolumn{6}{c}{Method} \\
    \cmidrule(lr){3-8}
      &  & GPI-LS & CAPQL & Q-Pensieve & MORL/D & \algoname-0 &\algoname \\
    \midrule
    \multirow{3}{*}{MO-Hopper-3d} 
      & HV\((10^7)\) & 5.78$\pm$5.49 & 4.53$\pm$2.64 & 5.23$\pm$3.89 & 8.97$\pm$0.09 &  10.20$\pm$0.36 & \textbf{10.63$\pm$0.40}\\
      & EU\((10^{2})\) & 1.63$\pm$0.11 & 2.51$\pm$0.79 & 2.49$\pm$0.88 & 3.66$\pm$0.21 & 3.58$\pm$0.35 & \textbf{3.69$\pm$0.23}\\
      & SP\((10^{2})\) & \textbf{0.23$\pm$0.09} & 29.04$\pm$16.92 & 48.44$\pm$37.93 & 11.64$\pm$3.24 & 1.33$\pm$0.98 & 1.35$\pm$1.14\\
    \midrule
    \multirow{3}{*}{MO-Ant-3d} 
      & HV\((10^7)\)  & 1.10$\pm$0.37 & 2.49$\pm$0.14 & 0.71$\pm$0.36 & 2.69$\pm$0.01 &  3.38$\pm$0.38 & \textbf{3.61$\pm$0.31} \\
      & EU\((10^{2})\)  & 1.31$\pm$0.38 & 1.90$\pm$0.03 & 1.09$\pm$0.25 & 1.95$\pm$0.04 & 1.89$\pm$0.14 & \textbf{1.96$\pm$0.15}\\
      & SP\((10^{2})\)  & 27.79$\pm$22.62 &3.72$\pm$3.70  & 51.57$\pm$51.73 & 10.38$\pm$15.85 & 0.65$\pm$0.28 & \textbf{0.54$\pm$0.15}\\
    \bottomrule
  \end{tabular}
  }

    \label{tab:metrics_3d}
\end{table*}

The experimental results, reported in Table \ref{tab:metrics_3d}, are highly consistent with the observations from our 2D benchmarks, showing that \algoname{} maintains an advantage in HV and EU over baselines. 
The robust performance in higher dimensions is a direct consequence of the algorithm's design, which generalises to $d$-objective tasks without requiring additional algorithmic modifications. As established in our theoretical analysis (Section \ref{subsec:theoretical}), the $d-1$ directional vectors obtained during the retraining phase are sufficient to explore and span the local Pareto manifold. This allows the locally linear extension mechanism to fill in contiguous patches of the Pareto surface effectively.

\subsection{Ablation Study}\label{subsec:ablation}
The \algoname{} integrates a locally linear extension process with a subsequent fine-tuning stage. To understand the distinct contributions of these components to the overall performance, our ablation study separates them. 
We first evaluate \algoname-0, which solely employs the extension process without fine-tuning. As detailed in Table \ref{tab:metrics_2d} and Table \ref{tab:metrics_3d}, 
\algoname-0 itself demonstrates competitiveness,
achieving strong HV and EU scores that are often competitive with or superior to baselines. This emphasises the efficacy of our extension mechanism in rapidly discovering a high-quality approximation of the Pareto front. 

Subsequently, we assess the improvement of the fine-tuning stage by comparing \algoname{} (which includes fine-tuning) to \algoname-0. 
This comparison reveals that the inclusion of fine-tuning yields further improvements in HV and EU across all settings. 
The impact on SP is less uniform, which is an expected outcome, as refining solutions towards a more optimal Pareto front can alter their relative spacing. 
Nevertheless, the consistent enhancements in HV and EU prove the value of fine-tuning for improving the quality of the approximate Pareto front and its coverage by diverse solutions. This demonstrates that the extension process provides a strong foundation for the fine-tuning stage that enables \algoname{} to outperform other algorithms.

\section{Related work}
\label{sec5:related_work}
Prior work in MORL offers various strategies for handling conflicting objectives. These can be broadly grouped into single-policy methods and multi-policy methods for approximating the Pareto front. 
Single-policy approaches typically convert the multi-objective problem into a single-objective task using a predefined preference or weighting scheme.
A common instance of such a weighting scheme is linear scalarization ~\citep{van2013scalarized}. Limitations of linear scalarization, particularly in capturing non-convex Pareto fronts, have been addressed by more advanced scalarization functions such as Chebyshev methods~\citep{van2013scalarized}, hypervolume-based approaches~\citep{zhang2020random} and the addition of concave terms to rewards~\citep{lu2023multi}.
Concurrently, significant efforts have developed generalised single-policy models conditioned on preference inputs to achieve adaptability across diverse objectives \citep{teh2017distral, yang2019generalized, basaklar2023pdmorl, parisi2016multi}, with subsequent extensions into offline learning contexts \citep{zhu2023scaling, lin2024policy} and methods to improve sample efficiency in these settings \citep{hung2023qpensieve}.

Multi-policy MORL strategies approximate the Pareto front by learning a diverse collection of policies.
One direction is developing single models conditioned on preferences~\citep{abels2019dynamic, friedman2018generalizing, kyriakis2022pareto}. 
Other approaches explicitly learn a diverse set of policies or their value functions, including direct value-based methods like Pareto Q-learning~\citep{van2014multi}, and evolutionary algorithms often guided by prediction models to discover a dense Pareto front~\citep{xu2020prediction}. Further techniques for generating policy sets involve Generalised Policy Improvement (GPI) for sample-efficient learning~\citep{alegre2023sample} or the development of transferable policy components using representations like successor features~\citep{alegre2022optimistic}. 
The use of constrained optimisation to efficiently complete and refine the Pareto front is also explored in~\citep{liu2024c, he2024personalized}.
Furthermore, the principles of decomposition-based strategies, which find a set of solutions by solving multiple interrelated scalarised sub-problems, have been a significant focus~\citep{felten2024multi, ropke2024divide}. 

While these single-policy and multi-policy paradigms have significantly advanced MORL, the characterisation and systematic exploitation of the structural relationship between the learned policies' underlying parameter space and their resultant performance on the Pareto front remain largely underexplored.
Existing studies have explored several related directions. 
In multi-objective optimisation, the Pareto-navigation method~\citep{ye2022pareto} uses objective-gradient information to search for preferred solutions within a Pareto set, while model-merging approaches~\citep{rame2023rewardedsoups,chen2025pareto} combine models optimised for different objectives to realise different trade-offs.
Within MORL, prior studies have touched on Pareto-set or policy-manifold learning, which aims to learn a global continuous representation of policies across preferences. This includes directly optimising a continuous policy manifold~\citep{parisi2016multi} and training a hypernetwork to map preferences to policy parameters~\citep{shu2024ijcai}. Related work also discusses parameter-space regularities~\citep{xu2020prediction}. 
Further studies characterise the exact geometry of Pareto fronts by analysing neighbouring deterministic policies~\citep{li2024find}, but only in known multi-objective MDPs that assume finite state and action spaces and known transition and reward models. Other studies investigate preference control~\citep{yuan2025moduli,yang2025preference}.

Although these directions provide related insights, they offer limited guidance on how to use local parameter-performance information for Pareto front reconstruction. More specifically, Pareto navigation, Pareto-set and policy-manifold learning remain primarily learning- or optimisation-based, treating the solution set as an object to be learned, represented, or optimised rather than exploiting its local parameter-performance geometry to guide reconstruction. Model-merging methods can combine discrete solutions, but this alone does not explicitly characterise the local Pareto-front manifold between them. Models independently optimised for different objectives do not by themselves establish how small parameter changes affect multi-objective performance, which is crucial for accurately tracing this manifold. In contrast, \algoname{} explicitly characterises the local parameter-performance relationship and reuses this structure around a sparse set of trained policies. Brief preference-based retraining provides local parameter-space directions that guide training-free policy-manifold tracing and Pareto-front reconstruction.

\section{Conclusion}
\label{sec6:conclusion}

We have discussed \algoname{}, 
an algorithm that identifies solution components in MORL by directly tracing local policy manifolds along the Pareto front.
The main benefit of \algoname{} is the structured representation it provides in policy parameter space, maintaining a direct relation to multi-objective performance, and thereby enabling increased efficiency in Pareto front approximation. 
We have shown, through both theoretical analysis and empirical evaluation, that this framework is sufficient to reconstruct a high-quality Pareto front, and that \algoname{} consistently outperforms recent MORL algorithms across a range of benchmark settings.
Furthermore, as the LLE stage is a modular plug-in component, it can be integrated into existing deep RL algorithms as a training-free and computationally lightweight module, offering a simple path for future extensions.


\begin{ack}

This work was supported by the United Kingdom Research and Innovation (grant EP/S023208/1), UKRI Centre for Doctoral Training in Robotics and Autonomous Systems at the University of Edinburgh, School of Informatics.
The authors sincerely thank Peize Li for providing additional computational resources to support the experiments, Haocheng Xiao for helpful discussions on figure presentation during the early stages of this work, and the anonymous reviewers for their insightful comments and suggestions.

\end{ack}

\bibliography{refs}
\bibliographystyle{plainnat}

\newpage
\appendix
\section{Algorithm}\label{app:algorithm}

\begin{algorithm}[h!]
\caption{\algoname{} (General $d$-Objective Case)}
\label{alg:pareto_extension}
\begin{algorithmic}[1]
\Require 
  Initial scalarization weights \(\{\omega_k\}_{k=1}^K\) evenly spanning the preference space, 
  Directional retraining weights $\{\omega_k^{(i)}\}_{k=1,i=1}^{K,m}$,
  Initialization training length \(T_{\mathrm{init}}\),
  Directional retraining length \(T_{\mathrm{dir}}\), Fine-tuning length \(T_{\mathrm{ref}}\), 
  Step-scale factors $\alpha_{\mathrm{start}},\alpha_{\mathrm{end}},\Delta\alpha$
\Ensure
  Approximate Pareto-optimal policy set \(\Pi\)

\State $\Pi \leftarrow \emptyset$
\State $m \leftarrow d-1$
\State $M \leftarrow \left\lfloor
\frac{\alpha_{\mathrm{end}}-\alpha_{\mathrm{start}}}{\Delta\alpha}\right\rfloor + 1$
\State
\State \textbf{Initialization:}
\For{\(k=1\) to \(K\)}
  \State Train base policy \(\theta_{\omega_k}\) with PPO \label{PPO_in_alg1} under weight \(\omega_k\) for \(T_{\mathrm{init}}\) steps
\EndFor
\State
\State \textbf{Directional Retraining:}
\For{\(k=1\) to \(K\)}
  \For{$i=1$ to $m$}
    \State $\theta_k^{(i)} \leftarrow$ continue training $\theta_{\omega_k}$ under $\omega_k^{(i)}$ for $T_{\mathrm{dir}}$ steps
    \State $\Delta\theta_k^{(i)} \leftarrow \theta_k^{(i)} - \theta_{\omega_k}$
    \State $\Delta \omega_k^{(i)} \leftarrow \omega_k^{(i)} - \omega_k$
  \EndFor
\EndFor
\State
\State \textbf{Locally Linear Extension:}
\State $\mathcal{C} \leftarrow \emptyset$
\For{$k=1$ to $K$}
  \ForAll{combination of $\alpha_i$ on the grid $[\alpha_{\mathrm{start}},\alpha_{\mathrm{end}}, \Delta\alpha]$}
    \State $\theta_{k,\mathrm{cand}} \leftarrow \theta_{\omega_k}+\sum_{i=1}^{m}\alpha_i\,\Delta\theta_k^{(i)}$
    \State $\omega_{k,\mathrm{cand}} \leftarrow \omega_k+ \sum_{i=1}^{m}\alpha_i\,\Delta \omega_k^{(i)}$
    \State Evaluate multi-objective performance $V(\theta_{k,\mathrm{cand}})$ under weight $\omega_{k,\mathrm{cand}}$
    \State $\mathcal{C} \leftarrow \mathcal{C} \cup \{(\theta_{k,\mathrm{cand}}, \omega_{k,\mathrm{cand}})\}$
  \EndFor
\EndFor
\State
\State \textbf{Candidate Selection:}
\State $\mathcal{N} \leftarrow$ non-dominated subset of $\mathcal{C}$
\State
  
\State \textbf{Preference-Aligned Fine-Tuning:}
\State $\mathcal{F} \leftarrow \emptyset$
\ForAll{$(\theta, \omega)\in \mathcal{N}$}
  \State fine‐tune $\theta$ for $T_{\mathrm{ref}}$ steps under $\omega$, yielding $\theta'$
  \State add $(\theta',\,\omega)$ to $\mathcal{F}$
\EndFor
\State $\mathcal{C}_{\text{all}} \leftarrow \mathcal{N} \,\cup\, \mathcal{F}$
\State $\mathcal{N}_{\text{final}} \leftarrow$ non‐dominated subset of $\mathcal{C}_{\text{all}}$\label{dominated_check}
\State $\Pi \leftarrow \Pi \;\cup\;\{\theta \mid (\theta,\cdot)\in \mathcal{N}_{\text{final}}\}$
\State
\State \textbf{return} $\Pi$

\end{algorithmic}
\end{algorithm}

In this section, we present a complete description of \algoname, an efficient procedure for tracing an approximate Pareto front in a
$d$-objective setting. Algorithm~\ref{alg:pareto_extension} details this process. 

The core pipeline involves several stages. 
First, \(K\) base policies \(\{\theta_{\omega_k}\}\) are trained, each under its respective initial weight \(\omega_k\) for \(T_{\mathrm{init}}\) steps. 
Next, for each base policy $\theta_{\omega_k}$, we perform directional retraining to obtain $m=d-1$ local directions: continuing training under nearby preferences $\omega_k^{(i)}$ for $T_{\mathrm{dir}}$ steps yields directional policies $\theta_k^{(i)}$, from which we compute the parameter updates $\Delta\theta_k^{(i)}\!=\!\theta_k^{(i)}-\theta_{\omega_k}$ and preference shifts $\Delta \omega_k^{(i)}\!=\!\omega_k^{(i)}-\omega_k$. 
In the Locally Linear Extension stage, we generate a candidate set $\mathcal{C}$ by combining these directions with step-scale factors $\{\alpha_i\}$ chosen from a grid $[\alpha_{\mathrm{start}},\alpha_{\mathrm{end}}]$ with step size $\Delta\alpha$, producing up to $M^m$ candidates per base policy together with matched weights $\omega_{k,\mathrm{cand}}$. 
We then select the non-dominated subset $\mathcal{N}\subset\mathcal{C}$ and apply Preference-Aligned Fine-Tuning for $T_{\mathrm{ref}}$ steps to obtain refined policies $\mathcal{F}$. 
Finally, \algoname{} returns $\Pi$, the non-dominated set extracted from $\mathcal{N}\cup\mathcal{F}$, which forms the final approximate Pareto front.

\section{Theoretical Analysis}\label{app:theory}
Throughout, $\|\cdot\|$ denotes the Euclidean norm on vectors by default, other norms on matrices are defined explicitly in the context.
\begin{definition}{Parameter-Performance Relationship (PPR).}\label{PPR_app}
Let \(\Theta\subseteq\mathbb{R}^n\) be the space of policy parameters, and $V$ be the expected discounted return as a function $V:\Theta\to\mathbb{R}^d$ of the parameter $\theta$. We say that $V$ exhibits a \emph{parameter–performance relationship} on an open region \(U\subseteq\Theta\) if $\exists\delta>0$ and a function $h:U\times B_\delta(0)\to\mathbb{R}^d$, \emph{s.t.} $\forall \theta\in U$ and perturbation \(\Delta\theta\) satisfying $\|\Delta\theta\|<\delta$ and \(\theta+\Delta\theta\in U\), the difference in the performance space is:
\begin{equation}
  V(\theta+\Delta\theta)\;-\;V(\theta)
  \;=\;h\bigl(\theta,\Delta\theta\bigr).
\end{equation}
\end{definition}

To clarify when LLE-MORL can reliably reconstruct a Pareto front, we present the theoretical analysis here. 
We consider here the Pareto fronts $P$ that are differentiable manifolds, meaning that they are locally similar to Euclidean space and derivatives can be defined.
As a slight generalisation, we can assume that the Pareto front is a non-connected manifold or a set of manifolds, including zero-dimensional manifolds (discrete points) or sets of manifolds of different dimensions. 
We will not further discuss this generalisation apart from mentioning the fact that it is possible to reach all components and to produce sufficiently many charts of each of the components of the Pareto front. 
For the theoretical framework, we propose the following assumptions:
\begin{assumption}{(Manifold Assumption)}\label{ass:manifold} The Pareto front $P$ is a set of differentiable manifold branches embedded in the performance space, which consists of all non-dominated points.
\end{assumption}
\begin{assumption}{(Continuity of PPR function $h$)}\label{ass:ppr}
The function $h$ in definition~\ref{PPR} is Lipschitz continuous with respect to the second variable $\Delta\theta$. 
\end{assumption}
\vspace*{-0.4cm}

\paragraph{Remark on Assumption~\ref{ass:ppr}:} This assumption can be reduced to more basic conditions on the expected discount return $V$, where $V$ is associated with the training phase through the PPO algorithm. Such a reduction is implemented by Theorem \ref{lc}.
\begin{assumption}{(Initialization)}\label{ass:initialization}
The initialisation in Algorithm \ref{alg:pareto_extension} stated in Section \ref{subsec:lle_algorithm} identifies $K$ different points on the optimal Pareto front. Thus, we have at least one accurate initial point for each branch of the Pareto front.
\end{assumption}

\begin{theorem}{(Lipschitz Continuity of PPR function $h$, corresponding to Theorem \ref{theo:lc})}\label{lc}
Denote the expected discounted return by $V:U\rightarrow\mathbb{R}^d$ where $U\subseteq\Theta\subseteq\mathbb{R}^n$ is an open set. The sufficient conditions for Lipschitz continuity of $h(\theta,\Delta\theta)$ with respect to the second variable $\Delta\theta$ are:\\[0.2cm]
(i) $V(\theta)$ is differentiable on $U$.\\
(ii) The score function is bounded, i.e. $\forall a\in\mathcal{A}, s\in\mathcal{S}$ 
\begin{equation}
\sup_{\theta\in U} \bigl\|\nabla_\theta \log \pi_\theta(a|s)\bigr\|\le \mathcal{G}<\infty.
\label{scorefunction}
\end{equation}
\end{theorem}
The inequality (ii) states that the score function \cite{baxter2001policy-gradient} is bounded by a constant $\mathcal{G}$. This is a normal regularity assumption in related research \cite{aydin2023policy-gradient,shi2024policy-gradient}.

\textit{\textbf{Proof}}: For a fixed $\theta\in U$, define $h_\theta(\Delta):=V(\theta+\Delta)-V(\theta)$. Suppose the objective return is:
\begin{equation}
V(\theta)=\mathbb{E}_{\pi_\theta}\!\left[\sum_{t=0}^{\infty}\gamma^t \boldsymbol{r}(s_t,a_t)\right]\in\mathbb{R}^d,
\end{equation}
with bounded rewards $\|\boldsymbol{r}(s,a)\|\le\mathcal{R}$, and $\pi_\theta(a|s)$ is a differentiable stochastic policy such that the score function $\nabla_\theta \log \pi_\theta(a|s)$ \cite{baxter2001policy-gradient} is locally bounded on a neighbourhood $U$ of $\theta$, i.e. Equation \ref{scorefunction}. Denote by $\tau_t=(s_0,a_0,\ldots,s_t,a_t)$
to be the trajectory up to time $t$, with density $p_\theta(\tau_t)$ on all possible trajectories $\mathcal T_t=(\mathcal S\times\mathcal A)^{t+1}$.

Take the gradient of $V$ about $\theta$ and make transformations:
\begin{align}
\nabla_\theta V(\theta)&=\nabla_\theta\mathbb{E}_{\pi_\theta}\!\left[\sum_{t=0}^{\infty}\gamma^t \boldsymbol{r}(s_t,a_t)\right]\\
&=\sum_{t=0}^{\infty}\gamma^t\nabla_\theta\mathbb{E}_{\pi_\theta}[\boldsymbol{r}(s_t,a_t)].
\label{eq:gradientV}
\end{align}
Further, the gradient of the reward expectation in Equation (\ref{eq:gradientV}) can be represented by
\begin{align}
\nabla_\theta\mathbb{E}_{\pi_\theta}[\boldsymbol{r}(s_t,a_t)]
&=\nabla_\theta\int_{\mathcal{T}_{t}}\!\boldsymbol{r}(s_t,a_t)p_\theta(\tau_t)\,d\tau_t\\
&=\int_{\mathcal{T}_{t}}\!\boldsymbol{r}(s_t,a_t)\cdot\nabla_\theta p_\theta(\tau_t)\,d\tau_t\\
&=\int_{\mathcal{T}_{t}}\!\boldsymbol{r}(s_t,a_t)p_\theta(\tau_t)\cdot\nabla_\theta\log p_\theta(\tau_t)\,d\tau_t\\
&=\mathbb{E}_{\pi_\theta}\left[\boldsymbol{r}(s_t,a_t)\cdot\nabla_\theta\log p_\theta(\tau_t)\right].
\label{eq:graditionE}
\end{align}
To handle the score function appears in Equation (\ref{eq:graditionE}), consider the joint distribution of trajectory $\tau_t$ 
\begin{align}
p_\theta(\tau_t)
&=\rho(s_0)
\prod_{k=0}^{t}\pi_\theta(a_k\mid s_k)
\prod_{k=0}^{t-1}P(s_{k+1}\mid a_k,s_k).
\label{eq:jointD}
\end{align}
Taking the logarithm of both sides of Equation (\ref{eq:jointD}) gives
\begin{align}
\nabla_\theta\log p_\theta(\tau_t)&=\nabla_\theta\log\rho(s_0)+\sum_{k=0}^{t}\nabla_\theta\log\pi_\theta(a_k\!\mid\! s_k)\\&+\sum_{k=0}^{t-1}
\nabla_\theta\log P(s_{k+1}\!\mid\! a_k,s_k)\\
&=\sum_{k=0}^{t}\nabla_\theta\log\pi_\theta(a_k\mid s_k).
\label{eq:gradientLog}
\end{align}
The first term is initial distribution and the third term is state transfer related to environment, they are not dependent on policy, thus has zero gradient.

Combine Equation (\ref{eq:gradientV},\ref{eq:graditionE},\ref{eq:gradientLog}) together, the gradient of $V$ is
\begin{align}
\nabla_\theta V(\theta)
&=\sum_{t=0}^{\infty}\gamma^t
\mathbb{E}_{\pi_\theta}\left[
\boldsymbol{r}(s_t,a_t)\cdot\nabla_\theta\log p_\theta(\tau_t)
\right]\\
&=\sum_{t=0}^{\infty}\gamma^t
\mathbb{E}_{\pi_\theta}\left[
\boldsymbol{r}(s_t,a_t)\cdot
\sum_{k=0}^{t}\nabla_\theta\log\pi_\theta(a_k\mid s_k)
\right]\\
&=\mathbb{E}_{\pi_\theta}\left[
\sum_{t=0}^{\infty}\gamma^t\boldsymbol{r}(s_t,a_t)\cdot
\sum_{k=0}^{t}\nabla_\theta\log\pi_\theta(a_k\mid s_k)
\right].
\label{eq:Vbound}
\end{align}
Above boundedness conditions on reward $\|\boldsymbol{r}\|\le\mathcal{R}$ and score function in Equation (\ref{scorefunction}) implies a uniform bound of $\nabla_\theta V$ by substituting in to Equation (\ref{eq:Vbound}):
\begin{equation}
\|\nabla_\theta V(\theta)\|
\le \sum_{t=0}^{\infty}\gamma^t \mathcal{R}(t+1)\mathcal{G}
= \frac{\mathcal{RG}}{(1-\gamma)^2}.
\end{equation}
Hence $V$ is locally Lipschitz on $U$. Further, let $L_V:=\sup_{\theta\in U}\|\nabla_\theta V(\theta)\|\leq \frac{\mathcal{RG}}{(1-\gamma)^2}$, we have:
\begin{equation}
\|h_\theta(\Delta_1)-h_\theta(\Delta_2)\|
= \|V(\theta+\Delta_1)-V(\theta+\Delta_2)\|
\le L_V \|\Delta_1-\Delta_2\|.
\label{eq1}
\end{equation}
Hence $h_\theta(\Delta)=h(\theta,\Delta)$ is Lipschitz on the ball $\Delta\in B_\delta(0)$ with $\theta+B_\delta(0)\subset U$:
In particular, Assumption \ref{ass:ppr} holds with Lipschitz constant $L_h\leq L_V\le R_{\max}G/(1-\gamma)^2$ under differentiability of $V$ and bounded score function conditions.\hfill$\square$


Theorem~\ref{lc} derives the Lipschitz continuity of PPR function $h$. Hence, small perturbations in the policy parameters only induce controllable small changes in the multi-objective performance vector $V$. In particular, along any path in the parameter space contained in the neighbour $\theta+B_\delta(0)$ of a fixed $\theta$, the image under $V:\Theta\rightarrow\mathbb{R}^d$ varies continuously and is bounded. This property is significant in reconstructing the Pareto surface produced by the LLE algorithm: with a suitable small step size, expanded policy candidates map to a controlled neighbourhood of the performance from initialisation. This guarantee is embodied and verified in the empirical results shown in Figure \ref{fig:moving}. Theorem \ref{lc} serves as one of the theoretical foundations that guarantees the performance of the central algorithm is stable in approximating and constructing the Pareto front.
\begin{corollary}{(Surface continuity guaranteed by Lipschitz PPR)}\label{cor:sep}
Algorithm LLE can construct a continuous surface in the performance space, as an approximation to the Pareto front, given appropriate regularity conditions and a range of $\alpha$.
\end{corollary}

\textit{\textbf{Proof}}: Suppose Theorem~\ref{lc} holds at a fixed $\theta$, i.e., $\exists$ $\delta, L_V>0$ such that for all $\Delta_1,\Delta_2$ with $\|\Delta_1\|,\|\Delta_2\|\le \delta$, we have: \[\|V(\theta+\Delta_1)-V(\theta+\Delta_2)\|\le L_V\|\Delta_1-\Delta_2\|.\]
For any two parameters $\theta_1,\theta_2\in \theta+B_\delta(0)$, if $\|V(\theta_1)-V(\theta_2)\|\ge \varepsilon>0$, we have $\|\theta_1-\theta_2\|\ge \varepsilon/L$. In particular, within the PPR neighbourhood $\theta+B_\delta(0)$, solutions that are separated obviously in the performance space must correspond to separated policies set in the parameter space, so LLE with sufficiently small step size cannot induce a spurious large displacement in performance space.\hfill$\square$

\begin{corollary}{(Lipschitz continuity of $V$)}\label{cor:V_lip} If Theorem \ref{lc} holds, then $V(\theta)$ is also Lipschitz continuous for $\theta\in U\subseteq\Theta$.
\end{corollary}

\textit{\textbf{Proof}}: $\forall{\theta_1,\theta_2}\in U\subseteq\Theta$, in Equation \ref{eq1}, substitute $\theta=\theta_2$, $\Delta_1=\theta_1-\theta_2$, $\Delta_2=0$. Then we immediately have:
\begin{equation}
    \|V(\theta_1)-V(\theta_2)\|\le L_V \|\theta_1-\theta_2\|.
\end{equation}
Thus, $V(\theta)$ is Lipschitz continuous with constant $L_V$.\hfill$\square$

\begin{condition}{(Chart of Pareto manifold)}\label{ass:branch}
Based on Assumption \ref{ass:manifold}, $\mathcal{P}\in \mathcal{M}^m(\mathbb{R}^d)$, where the dimension of $\mathcal{P}$ is $m=d-1$, and $\forall v\in\mathcal{P}$ is non-dominated. By definition, $\forall v\in\mathcal{P}$, there exists an open neighbourhood $P_v\subset\mathcal{P}$ and a chart:
\begin{equation}
    \varphi_v: P_v\rightarrow\Omega_v\subset\mathbb{R}^{d-1}
\end{equation}
where $\Omega_v$ is an open set. Without loss of generality, we choose the chart such that $\varphi_v(v)=0$. And $\varphi_v$ is a $C^k$ diffeomorphism ($k\geq1$), with the inverse mapping is denoted by: $\varphi_v^{-1}:\Omega_v\rightarrow P_v\subset\mathbb{R}^d$.
\end{condition}

\begin{condition}{(Local lifting of a Pareto neighbourhood)}\label{ass:parambranch_hd} 
Based on Condition \ref{ass:branch}, on the Pareto front, consider a fixed $v\in\mathcal{P}$ and its open neighbourhood $P_v\subset\mathcal{P}$. Suppose a corresponding policy parameter $\theta_v$ satisfying $V(\theta_v)=v$. Define an equivalence relation $\sim$ on $\Theta$ by $\theta_1\sim\theta_2$ if $V(\theta_1)=V(\theta_2)$, and denote the quotient space by $\Theta/{\sim}$. Assume there exist $\Theta_v\subset\Theta$ consisting of the representative elements of $\Theta/{\sim}$, satisfying the following conditions:
\begin{flalign}
&(i)\ \exists{\delta}>0,\ s.t.\ \Theta_v\subseteq B_\delta(\theta_v).&&\\
&(ii)\ V[\Theta_v]= P_v.\\
&(iii)\ \forall \theta\in\Theta_v,\ |\Theta_v\cap{[\theta]}|=1.
\end{flalign}
where $B_{\delta}(\theta_v)$ denote an open ball with radius $\delta$ centres on $\theta_v$, $[\theta]\in\Theta/{\sim}$ represent the equivalence class of $\theta$, and $|\cdot|$ denote the cardinality of set. By $(ii)$, $V|_{\Theta_v}\!:\!\Theta_v\twoheadrightarrow P_v$ is a surjection. By $(iii)$, we have $V|_{\Theta_v}\!:\!\Theta_v\hookrightarrow P_v$ is an injection. Therefore, $V|_{\Theta_v}\!:\!\Theta_v{\rightarrow}P_v$ is a bijection. Assume further that $\theta_v\in\Theta_v\subset U$ and $(V|_{\Theta_v})^{-1}\!:\!P_v\to\Theta_v\subset\mathbb{R}^n$ is a $C^k$ map ($k\ge 2$).
\end{condition}

Condition~\ref{ass:parambranch_hd} serves two purposes. Firstly, the localisation $\Theta_v\subset B_\delta(\theta_v)$ and $V(\Theta_v)=P_v$ ensures that the entire Pareto neighbourhood $P_v$ can be lifted to a region in parameter space, so the algorithm does not need to traverse distant solution components. This is the prerequisite under which local linear extrapolation in parameter space is meaningful for approximating the same Pareto branch. Second, the representative constraint $|\Theta_v\cap[\theta]|=1$ removes the multi-to-one ambiguity of $V$ by selecting exactly one representative from each equivalence class, so that notions such as parameter proximity, difference directions, and extrapolation updates are well-defined on the lifted set. More specifically, it formulates an injection from $\Theta_v$ to the Pareto neighbourhood of $v$. Without this condition, a single performance could correspond to many distant parameters and retrained $\Delta\theta$ directions would be unstable and unreliable in exploring the Pareto front.
\begin{figure}[ht]
    \centering
    \includegraphics[width=0.7\linewidth]{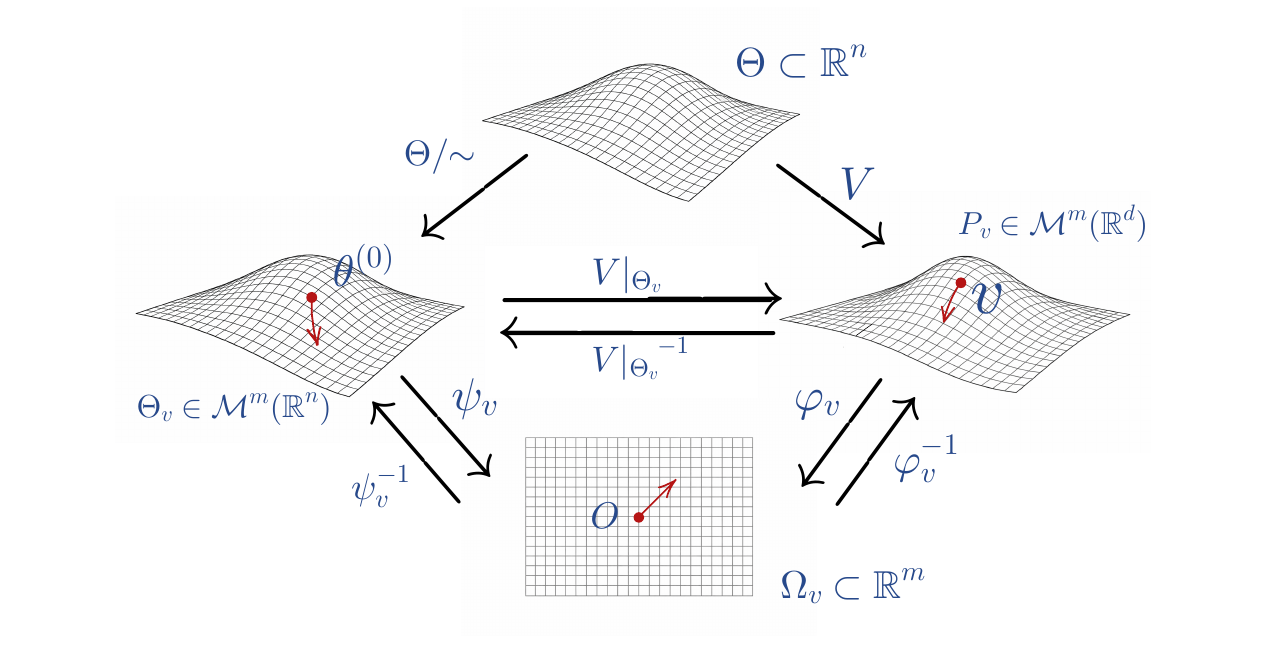}
    \caption{Commutative diagram between Pareto front, parameter manifold and their local lifting. With each initialisation $\theta^{(0)}$, we denote its performance $v$ and the neighbour on the Pareto front manifold to be $P_v$. Based on our assumptions and conditions, the local lifting, i.e. $\Theta_v$ is also a manifold of the same dimension as $P_v$, located in the neighbourhood of $\theta^{(0)}$. The Linear extrapolation corresponds to the approximation of the tangent space of $\Theta_v$. Each time, \algoname{} reconstructs the Pareto front branch where $v$ is located. With the appropriate choices of initialization $\theta_k^{(0)}$, it reconstructs all of the branches where the corresponding $v_k=V\Big(\theta_k^{(0)}\Big)$ is located.}
    \label{fig:placeholder}
\end{figure}
\begin{theorem}{(Local lifting manifold, w.r.t. Theorem \ref{theo:thm:smooth_lift})}\label{thm:smooth_lift}
Let $P_v\!\subset\!\mathcal{P}$ be as in Condition~\ref{ass:branch}, where $P_v\in\mathcal{M}^m(\mathbb{R}^d)$ with a $C^k$ chart $\varphi_v:P_v\to\Omega_v\subset\mathbb{R}^m$. Suppose $V|_{\Theta_v}\!:\!\Theta_v\to P_v$ is a bijection, $V$ is differentiable on $U$ and
$(V|_{\Theta_v})^{-1}:P_v\to\mathbb{R}^n$ is $C^k$
as in Condition~\ref{ass:parambranch_hd}. Then we have $\Theta_v$ defined in Condition \ref{ass:parambranch_hd} is also a differential manifold $\mathcal{M}^m(\mathbb{R}^n)$, with an $C^k$ chart $\psi_v:\Theta_v\rightarrow\Omega_v\subset\mathbb{R}^m$.
\end{theorem}
\textit{\textbf{Proof}}: Fix a $C^k$ chart $\varphi_v:P_v\!\to\!\Omega_v\subset\mathbb{R}^m$ from Condition~\ref{ass:branch}. Define $\psi_v:=\varphi_v\circ V|_{\Theta_v}:\Theta_v\to\Omega_v$
and $\psi_v^{-1}:=(V|_{\Theta_v})^{-1}\circ\varphi_v^{-1}:
\Omega_v\to\mathbb{R}^n$.
By conditions on $(V|_{\Theta_v})^{-1}$ and $\varphi_v^{-1}$, $\psi_v^{-1}$ is also $C^k$ and injective. It follows $\psi_v$ is continuous since $V|_{\Theta_v}$ and $\varphi_v$ are continuous. 

Consider the composition mapping $V|_{\Theta_v}\circ\psi_v^{-1}=\varphi_v^{-1}$, chain rule gives
\begin{equation}
 D V\Big(\psi_v^{-1}(x)\Big)\cdot  D\psi_v^{-1}(x)
=D\varphi_v^{-1}(x).
\end{equation}
As $\varphi_v^{-1}$ is a local parametrisation of an
$m$-dimensional embedded manifold by Condition~\ref{ass:branch},
it follows that $\mathrm{rank}\big(D\varphi_v^{-1}(x)\big)=m$. By the rank inequality for matrix products,
\begin{equation}
\begin{aligned}
m=\mathrm{rank}\big(D\varphi_v^{-1}(x)\big)\leq\mathrm{rank}\big(D\psi_v^{-1}(x)\big)\leq m,
\end{aligned}
\end{equation}
where the last inequality follows from
$D\psi_v^{-1}(x)\in\mathbb{R}^{n\times m}$.
Therefore,
$\mathrm{rank}\big(D\psi_v^{-1}(x)\big)=m$.\\
Hence $\psi_v^{-1}$ is an injective immersion with
continuous inverse $\psi_v$, and therefore a $C^k$
embedding. Consequently, $\Theta_v$ is an
$m$-dimensional $C^k$ manifold embedded in
$\mathbb{R}^n$, i.e.
$\Theta_v\in\mathcal{M}^m(\mathbb{R}^n)$. $\psi_v$ is a $C^k$ diffeomorphism between
$\Theta_v$ and the open set $\Omega_v\subset\mathbb{R}^m$.\hfill$\square$

\begin{condition}{(Local extension directions, corresponding to Condition \ref{theo:con:directions_hd})}\label{ass:directions_hd}
Fix $v\in\mathcal{P}$, an open neighbourhood $P_v\subset\mathcal{P}$ as in Condition \ref{ass:branch}, and the corresponding lifted set $\Theta_v\subset\Theta$ from Condition~\ref{ass:parambranch_hd}. Suppose the initialization returns $\theta^{(0)}\in\Theta_v$ with $V(\theta^{(0)})=v$ by Assmption \ref{ass:initialization}.
Applying $m=\dim\mathcal{P}$ distinct nearby preference perturbations and directional retraining from $\theta^{(0)}$, one obtains
$\theta^{(i)}\in\Theta_v$, $i=1,\dots,m$, such that
\begin{equation}
\mathcal{D}:=\bigl[\theta^{(1)}-\theta^{(0)},\ \theta^{(2)}-\theta^{(0)},\ \dots,\ \theta^{(m)}-\theta^{(0)}\bigr]\in\mathbb{R}^{n\times m}
\end{equation}
has full column rank, i.e. $\mathrm{rank}(\mathcal{D})=m$. The approximation of these finite-difference directions by tangent vectors is quantified in Lemma~\ref{lem:X_inv}. 
This explains Step (2) in the Algorithm \ref{alg:pareto_extension} in Section \ref{subsec:lle_algorithm}.
\end{condition}

\begin{lemma}{(Coordinates of linear extension and error matrix)}\label{lem:X_inv}
Let $\psi_v$ be as in Theorem~\ref{thm:smooth_lift}, without loss of generality, define $\psi_v(\theta^{(0)})=0$, and denote $x^{(i)}:=\psi_v(\theta^{(i)})$ and $X:=[x^{(1)},\dots,x^{(m)}]$. Let $\mathcal{J}:=D\psi_v^{-1}(0)$ be the Jacobian at the origin of $\psi_v^{-1}$, and suppose $\psi_v^{-1}$ is $C^2$ with $\|D^2(\psi_v^{-1})(x)\|\le \eta$ on a convex neighbourhood of $0$ contained in $\Omega_v$ and containing all $x^{(i)}$, $i=1,\dots,m$. 

Define $\mathcal{D}$ as in Condition~\ref{ass:directions_hd}. Then there exists error matrix $E\in\mathbb{R}^{n\times m}$ such that: $\mathcal{D}=\mathcal{J}X+E$ and its $L_2$ norm is controlled by:
\begin{equation}
    \|E\|_2\le \frac{\eta}{2}\sqrt{m}\,\max_{1\le i\le m}\|x^{(i)}\|^2.
    \label{eq:errormatrix}
\end{equation}
Let $\sigma_{\min}(\mathcal{D})$ be the smallest singular value of $\mathcal{D}$. 
If $\|E\|_2<\sigma_{\min}(\mathcal{D})$, then $X$ is invertible.
\end{lemma}

\textit{\textbf{proof}}: (I). Apply Taylor expansion of $\psi_v^{-1}$ at $0$, where $\psi_v^{-1}(0)=\theta^{(0)}$:
\begin{equation}
    \psi_v^{-1}(x^{(i)})=\psi_v^{-1}(0)+\mathcal{J}x^{(i)}+r^{(i)}
\end{equation}
where $r^{(i)}$ are remainder of the second order. Equivalently,
\begin{equation}
    \theta^{(i)}-\theta^{(0)}=\mathcal{J}x^{(i)}+r^{(i)}
\end{equation}
Arrange all formulas of $1\leq i\leq m$ in matrix form, we have:
\begin{equation}
    \mathcal{D}=\mathcal{J}X+E
\end{equation}
Matrix $E:=\left[r^{(1)},r^{(2)},\dots,r^{(m)}\right]\in\mathbb{R}^{n\times m}$ is named the error matrix as it gives the error of the linear approximation. Matrix $X$ collects the chart coordinates of the retraining points $\theta^{(1)},\dots,\theta^{(m)}$.

Next, we derive the upper bound of $E$. Denote the components of $\psi_v^{-1}=\left[\psi^{-1}_1,\dots,\psi^{-1}_n\right]$ with $\psi_\ell^{-1}:\Omega_v\subset\mathbb{R}^m\to\mathbb{R}$.
For $x\in\Omega_v$ and $u,w\in\mathbb{R}^m$, define the second derivative as a bilinear map
\begin{equation}
D^2(\psi_v^{-1})(x)[u,w]:=\left[u^\top \nabla^2\psi^{-1}_1(x)\,w,\ \dots,\ u^\top \nabla^2\psi^{-1}_n(x)\,w\right]^\top \in\mathbb{R}^n,
\end{equation}
where $\nabla^2\psi^{-1}_\ell(x)=\left[\frac{\partial^2\psi^{-1}_\ell}{\partial x_i\partial x_j}(x)\right]_{i,j=1}^m$ is the Hessian of $\psi_\ell^{-1}$. Accordingly, define the operator norm: \begin{equation}
\|D^2(\psi_v^{-1})(x)\|:=\sup_{u,w\neq 0}\frac{\|D^2(\psi_v^{-1})(x)[u,w]\|_2}{\|u\|\cdot\|w\|}
\end{equation}

By Taylor's theorem, with the notation above, the reminder of $\psi_v^{-1}$, i.e. $r^{(i)}$ has an integral form:
\begin{equation}
    \psi_v^{-1}(x^{(i)})=\psi_v^{-1}(0)+\mathcal{J}x^{(i)}+\int_0^1(1-t)D^2(\psi_v^{-1})(tx^{(i)})[x^{(i)},x^{(i)}]dt.
\end{equation}
Thus,
\begin{align}
    \|r^{(i)}\|&\le \int_0^1(1-t)\,\|D^2(\psi_v^{-1})(tx^{(i)})[x^{(i)},x^{(i)}]\|\,dt\\
    &\le\int_0^1(1-t)\,\eta\|x^{(i)}\|^2\,dt\\
    &= \frac{\eta}{2}\|x^{(i)}\|^2.
    \label{eq:r}
\end{align}
Then we can control the norm of $E$, i.e. Equation \ref{eq:errormatrix}, by the upper bound of its Frobenius norm:
\begin{equation}
    \|E\|_2\le \|E\|_F=\left(\sum_{i=1}^m\|r^{(i)}\|^2\right)^{\frac{1}{2}}\le \frac{\eta}{2}\left(\sum_{i=1}^m\|x^{(i)}\|^4\right)^{\frac{1}{2}}\le \frac{\eta}{2}\sqrt m\,\max_{1\leq i\leq m}\|x^{(i)}\|^2.
\end{equation}
(II). If $X$ were singular, there would exist $u\in\mathbb{R}^m,\ u\neq0$ with $Xu=0$, hence $\mathcal{D}u=Eu$. Therefore
\begin{equation}
\|\mathcal{D}u\|\le \|E\|_2\cdot\|u\|.
\end{equation}
On the other hand, by definition of $\sigma_{\min}(\mathcal{D})$,
\begin{equation}
\|\mathcal{D}u\|=\|u\|\cdot\left\|\mathcal{D}\frac{u}{\|u\|}\right\|\ge \|u\|\cdot \inf\limits_{\|x\|=1}\|\mathcal{D}x\|_2   \ge\sigma_{\min}(\mathcal{D})\,\|u\|.
\end{equation}
Thus $\sigma_{\min}(\mathcal{D})\le \|E\|_2$, contradicting $\|E\|_2<\sigma_{\min}(\mathcal{D})$. Therefore $X$ is invertible. Furthermore, for every $u\in\mathbb R^m$, there is
\[
\|\mathcal D u-\mathcal JXu\|
=\|Eu\|
\le\|E\|_2\|u\|
\le
\frac{\|E\|_2}{\sigma_{\min}(\mathcal D)}
\|\mathcal D u\|.
\]
Since $T_{\theta^{(0)}}\Theta_v=\operatorname{range}(\mathcal J)$,
we have $\mathcal JXu\in T_{\theta^{(0)}}\Theta_v$.
Thus, the relative approximation error is small when
$\|E\|_2/\sigma_{\min}(\mathcal D)$ is small.
\hfill$\square$

All above, the preparation work is finished. We now focus on the core theorem supporting the effectiveness of the LLE algorithm. Notations in previous deductions are continuously adopted.
\begin{theorem}{(Local Pareto error of LLE, corresponding to Theorem \ref{theo:theorem:lle})}\label{theorem:lle} Denote the Linear exploration by $\tilde\theta(\alpha):=\theta^{(0)}+\mathcal{D}\alpha$. Then for a suitable range of $\alpha\in\mathbb{R}^m$ with $X\alpha\in\Omega_v$, the error of exploration to the actual Pareto front is controlled by:
\begin{equation}
\mathrm{dist}\bigl(V(\tilde\theta(\alpha)),\,\mathcal{P}_v\bigr)
\le \frac{L_V \eta}{2}\max_{1\leq i\leq m}\|x^{(i)}\|^2\Big(\sqrt m\,\|\alpha\|+m\,\|\alpha\|^2\Big).
\label{eq:error_bound}
\end{equation}
\end{theorem}
\textit{\textbf{Proof}}:
Define the point-to-set distance in performance space by
\begin{equation}
\mathrm{dist}(z,\mathcal{P}_v):=\inf_{p\in\mathcal{P}_v}\|z-p\|,\qquad z\in\mathbb{R}^d.
\end{equation}
Let $\psi_v:\Theta_v\to\Omega_v$ be the $C^2$ chart from Theorem~\ref{thm:smooth_lift} with $\psi_v(\theta^{(0)})=0$ and
set $\theta^\star(\alpha):=\psi_v^{-1}(X\alpha)\in\Theta_v$, which is the lifted manifold point with chart coordinate $X\alpha$. Since $V[\Theta_v]=\mathcal{P}_v$, we have
$V(\theta^\star(\alpha))\in\mathcal{P}_v$ whenever $X\alpha\in\Omega_v$.
Hence,
\begin{equation}
\mathrm{dist}\bigl(V(\tilde\theta(\alpha)),\mathcal{P}_v\bigr)
\le \bigl\|V(\tilde\theta(\alpha))-V(\theta^\star(\alpha))\bigr\|_2.
\label{eq:dist}
\end{equation}
By the local Lipschitz continuity of $V$ on the neighbourhood under consideration,
\begin{equation}
\bigl\|V(\tilde\theta(\alpha))-V(\theta^\star(\alpha))\bigr\|_2
\le L_V\,\bigl\|\tilde\theta(\alpha)-\theta^\star(\alpha)\bigr\|_2.
\label{eq:V}
\end{equation}
It remains to bound $\|\tilde\theta(\alpha)-\theta^\star(\alpha)\|_2$. By Lemma~\ref{lem:X_inv}, for the retraining points
$\theta^{(i)}\in\Theta_v$ we have $\mathcal{D}=\mathcal{J}X+E$ with $\mathcal{J}:=D(\psi_v^{-1})(0)$.
By Taylor's theorem with integral remainder applied to $\psi_v^{-1}$,
\begin{equation}
\psi_v^{-1}(X\alpha)=\psi_v^{-1}(0)+\mathcal{J}(X\alpha)+r(X\alpha)
\end{equation}
By Equation \ref{eq:r}, $\|r(X\alpha)\|\le \frac{\eta}{2}\|X\alpha\|^2$. Using $\psi_v^{-1}(0)=\theta^{(0)}$ and $\tilde\theta(\alpha)=\theta^{(0)}+\mathcal{D}\alpha$, we obtain
\begin{align}
\tilde\theta(\alpha)-\psi_v^{-1}(X\alpha)&=\left(\theta^{(0)}+\mathcal{D}\alpha\right)-\left(\theta^{(0)}+\mathcal{J}X\alpha+r(X\alpha)\right)\\
&=E\alpha-r(X\alpha).
\end{align}
Therefore,
\begin{equation}
\bigl\|\tilde\theta(\alpha)-\theta^\star(\alpha)\bigr\|
=\bigl\|E\alpha-r(X\alpha)\bigr\|
\le \|E\|_2\,\|\alpha\|+\frac{\eta}{2}\|X\alpha\|^2.
\label{eq:theta}
\end{equation}
Combining inequalities of Equation \ref{eq:theta}, \ref{eq:dist}, \ref{eq:V}, yields:
\begin{align}
\mathrm{dist}\bigl(V(\tilde\theta(\alpha)),\mathcal{P}_v\bigr)\le L_V\Big(\|E\|_2\cdot\|\alpha\|+\frac{\eta}{2}\|X\alpha\|^2\Big)
\label{eq:dist2}
\end{align}
Further, let $\|\cdot\|_F$ be Frobenius norm,  
\begin{align}\|X\alpha\|^2 &\le\|X\|_2^2\cdot\|\alpha\|^2\\&\le \|X\|_F^2\cdot\|\alpha\|^2\\&= \|\alpha\|^2\sum_{i=1}^m \|x^{(i)}\|^2\\&\le m\|\alpha\|^2\max_{1\leq i\leq m}\|x^{(i)}\|^2
\label{eq:Xalpha}
\end{align}
Substitute Equation \ref{eq:errormatrix} from Lemma \ref{lem:X_inv}, Equation \ref{eq:Xalpha} into Equation \ref{eq:dist2}, we have :
\begin{equation}
\mathrm{dist}\bigl(V(\tilde\theta(\alpha)),\mathcal P_v\bigr)\le \frac{L_V \eta}{2}\max_{1\leq i\leq m}\|x^{(i)}\|^2\Big(\sqrt m\,\|\alpha\|+m\,\|\alpha\|^2\Big),
\end{equation}
which completes the proof.\hfill$\square$

Under the conditions of the theorem, we describe the LLE-MORL time complexity as follows.
\begin{proposition}\label{proposition:time} (Time complexity)
Given that the number of objectives is $n$, the number of base policies is $K$, the base policy training time is $T$, the number of locally linear extensions sampled per direction is $M$, and noting that the locally linear extension is training-free, the expected running time of LLE-MORL is $O(TK+KM^{n-1})$.
\end{proposition}

\textit{Proof for proposition \ref{proposition:time}.}
In an $n$-objective reinforcement learning problem, if the Pareto front in the policy parameter space can be represented as a differentiable $(n-1)$-dimensional manifold, LLE-MORL can locally reconstruct this manifold by extending each base policy along $(n-1)$ distinct directions obtained via directional retraining.
The locally linear extension then combines these directions to generate a grid of new candidate policies. In a $n$-objective task, $n-1$ directional retrained directions are used to form a $(n-1)$-dimensional grid: $\theta=\theta_{\rm base}+\sum\alpha_i(\theta_{i}-\theta_{\rm base})$, $i=1,\dots,n-1$, allowing dense coverage of the local Pareto surface with or without further training. 
 Training $K$ base policies requires $O(TK)$. Around each base policy, local linear extension samples $M^{n-1}$ candidate policies across the $n-1$-dimensional tangent space. These extensions require no retraining, so their cost is dominated by evaluation, yielding the second term $O(KM^{n-1})$. Thus the overall runtime is $O(TK+KM^{n-1})$.

The first term in the time complexity proposition corresponds to the cost of training $K$ base policies, which is the dominant cost. The second term accounts for generating and evaluating $M^{n-1}$ extended policies per base policy. This step is training-free and its cost is controllable via $M=(\frac{\alpha_{\rm max}-\alpha_{\rm min}}{\Delta\alpha})$, allowing the user to balance Pareto front accuracy/density against computational resources. The second term grows exponentially with the number of objectives, but for low- to moderate-dimensional objectives, this cost remains small compared to policy training. For very high-dimensional objective spaces, adaptive sampling or sparse grids can mitigate this cost. 

\section{Experiment Setup Details}\label{app:exp_details}

\subsection{Benchmarks} 

To evaluate the performance of our proposed \algoname{} method and compare it against existing baselines, we utilise a suite of continuous control benchmarks from the MO-Gymnasium library~\citep{felten_toolkit_2023}. These environments are designed to test the ability of an agent to learn policies that effectively balance multiple, often conflicting objectives. The specific environments and their multi-objective reward formulations are detailed below:

\paragraph{\textbf{\textsc{MO-Swimmer-v5.}}}
A planar, three-link swimmer operating in a viscous fluid, utilising a 2D continuous action space to control its joint torques. The objectives are to maximise forward velocity along the $x$-axis and minimise the control cost.

The observation space \(\mathcal{S} \subset \mathbb{R}^8\) includes joint angles and velocities, and the action space \(\mathcal{A} \subset \mathbb{R}^2\) represents joint torques in \([-1, 1]\).
Let \(x_{\text{before}}\) and \(x_{\text{after}}\) be the $x$-coordinates of the centre of mass of the swimmer before and after an action, \(\Delta t\) be the time step, and \(a_j\) be the $j$-th component of the action vector. 

The first objective is the forward speed
\[
R_1 = \frac{x_{\text{after}} - x_{\text{before}}}{\Delta t},
\]
and the second objective is the energy efficiency (negative control cost):
\[
R_2 = -  \sum_{j} a_j^2 
\]

\paragraph{\textbf{\textsc{MO-Hopper-2obj-v5.}}}
This environment features a 2D one-legged hopper with a 3-dimensional continuous action space controlling torques for its thigh, leg, and foot joints. Originally a 3-objective task (forward speed, jump height, control cost), we use the 2-objective variant, in which the separate control-cost objective is added to other objectives.

The observation space \(\mathcal{S} \subset \mathbb{R}^{11}\) includes joint states and torso position, and the action space \(\mathcal{A} \subset \mathbb{R}^3\) represents joint torques in \([-1, 1]\).
Let \(v_x=(x_{\text{after}}-x_{\text{before}})/{\Delta t}\) be the forward velocity of the agent along the $x$-axis, where \(x_{\text{after}}\) and \(x_{\text{before}}\)  are x-positions of the torso.
Let \(h_{\text{jump}}= 10 \times (z_{\text{after}} - z_{\text{init}})\) be a measure of jumping height, where \(z_{\text{after}}\) is the current z-position of the torso and \(z_{\text{init}}\) is its initial z-position.
Let \(c_{\text{ctrl}}\) be the positive control cost, computed as \(w_{\text{env\_ctrl}} \sum_{j} (a_j)^2\), where \(w_{\text{env\_ctrl}}\) is the environment control cost weight (typically \(0.001\)).
Let \(r_{\text{healthy}}\) be the health reward (typically \(+1\) if the agent has not fallen).
The reward vector \(\mathbf{R} = [R_1, R_2]\) is defined as:
\begin{itemize}
    \item \(R_1\) (Adjusted Forward Performance):
    \[
    R_1 = v_x + r_{\text{healthy}}- c_{\text{ctrl}} 
    \]
    \item \(R_2\) (Adjusted Height Performance):
    \[
    R_2 = h_{\text{jump}} + r_{\text{healthy}} - c_{\text{ctrl}} 
    \]
\end{itemize}

\paragraph{\textbf{\textsc{MO-Ant-2obj-v5.}}}
A quadrupedal “ant” robot in 2D with an eight-dimensional action space for joint torques. By default, the environment emits a three-dimensional reward vector: (1) \(x\)-velocity, (2) \(y\)-velocity, and (3) control cost.  Here, we use the two-objective variant in which the separate control-cost objective is added to other objectives.

The observation space \(\mathcal{S} \subset \mathbb{R}^{27}\) includes joint states, torso position, and contact forces, and the action space \(\mathcal{A} \subset \mathbb{R}^8\) represents joint torques in \([-1, 1]\).
Let \(v_x=(x_{\text{after}}-x_{\text{before}})/{\Delta t}\) be the forward velocity of the agent along the $x$-axis, where \(x_{\text{after}}\) and \(x_{\text{before}}\)  are x-positions of the torso.
Let \(v_y=(y_{\text{after}}-y_{\text{before}})/{\Delta t}\) be the forward velocity of the agent along the $y$-axis, where \(y_{\text{after}}\) and \(y_{\text{before}}\)  are y-positions of the torso.
Let \(c_{\text{ctrl}}\) be the positive control cost, computed as \(w_{\text{env\_ctrl}} \sum_{j} (a_j)^2\), where \(w_{\text{env\_ctrl}}\) is the environment control cost weight (typically 0.05).
Let \(r_{\text{healthy}}\) be the health reward (typically \(+1\) if the Ant is healthy).
Let \(p_{\text{contact}}\) be the positive contact penalty, which is used for penalising the Ant if the external contact forces are too large, computed as \(w_{\text{env\_contact}} \sum_{k} (\text{force}_k)^2\), where \(w_{\text{env\_contact}}\) is the environment contact cost weight (typically \(5 \times 10^{-4}\)).
The reward vector \(\mathbf{R} = [R_1, R_2]\) is defined as:
\begin{itemize}
    \item \(R_1\) (Adjusted $x$-Velocity Performance):
    \[
    R_1 = v_x  + r_{\text{healthy}} - c_{\text{ctrl}} - p_{\text{contact}}
    \]
    \item \(R_2\) (Adjusted $y$-Velocity Performance):
    \[
    R_2 = v_y+ r_{\text{healthy}} - c_{\text{ctrl}}  - p_{\text{contact}}
    \]
\end{itemize}

\paragraph{\textbf{\textsc{MO-Hopper-3obj-v5.}}}
This environment is identical to \textsc{MO-Hopper-2obj-v5} in terms of state and action spaces, but retains all three original objectives instead of merging the control cost into the other rewards. As a result, the task features three objectives: forward speed, jump height, and control cost.
The reward vector \(\mathbf{R} = [R_1, R_2,R_3]\) is defined as:
\begin{itemize}
    \item \(R_1\) ($x$-Velocity Performance):
            \[R_1 = v_x + r_{\text{healthy}}\]
    \item \(R_2\) ($y$-Velocity Performance):
            \[R_2 = h_{\text{jump}} + r_{\text{healthy}}\]
    \item \(R_3\) (Control Cost):
            \[R_3 = -\,c_{\text{ctrl}} + r_{\text{healthy}}\]
\end{itemize}

\paragraph{\textbf{\textsc{MO-Ant-3obj-v5.}}}
This environment is identical to \textsc{MO-Ant-2obj-v5} in terms of state and action spaces, but retains all three original objectives instead of merging the control cost into the other rewards. As a result, the task features three objectives: $x$-velocity, $y$-velocity, and control cost.
The reward vector $\mathbf{R} = [R_1, R_2, R_3]$ is defined as:
\begin{itemize}
    \item $R_1$ ($x$-Velocity Performance):
    \[
    R_1 = v_x + r_{\text{healthy}} - p_{\text{contact}}
    \]
    \item $R_2$ ($y$-Velocity Performance):
    \[
    R_2 = v_y + r_{\text{healthy}} - p_{\text{contact}}
    \]
    \item $R_3$ (Control Cost):
    \[
    R_3 = -\,c_{\text{ctrl}}
          + r_{\text{healthy}}
          - p_{\text{contact}}
    \]
\end{itemize}

\subsection{Evaluation Metrics}

We evaluate the quality of the approximate Pareto front using three standard metrics, following the formalism in~\citep{zitzler2002multiobjective, zintgraf2015quality, hayes2022practical}.

{\bf Hypervolume (HV).}  
Let \(P\) be an approximate Pareto front and \(G_0\) a reference point dominated by all \(p\in P\).  The hypervolume is
\(
  \mathcal{H}(P)
  = \int_{\mathbb{R}^d} \mathbbm{1}_{H(P)}(z)\,dz,
\)
where
\(
  H(P) = \{\,z \in \mathbb{R}^n \mid \exists\,j,\;1\le j\le |P|\;:\;G_0 \preceq z \preceq P(j)\}
\). 
Here, \(P(j)\) is the \(j^{\rm th}\) solution in \(P\), the symbol \(\preceq\) denotes objective dominance, and \(\mathbbm{1}_{H(P)}\) is an indicator 
function that equals 1 if \(z\in H(P)\) and 0 otherwise. A higher hypervolume implies a front closer to and more extensive with respect to the true Pareto front.

{\bf Expected Utility (EU).}  
Let \(P\) be an approximate Pareto front and \(\Pi\) be the corresponding policy set. The expected utility metric is
\(
  \mathcal{U}(P)
  = \mathbb{E}_{\omega\sim\Omega}\bigl[\max_{\pi\in\Pi}\;\omega^\top G^\pi_\omega\bigr].
\)
A higher EU denotes better average performance over preferences.

{\bf Sparsity (SP).}  
Let \(P\) be an approximate Pareto front in a \(d\)-dimensional objective space. The sparsity metric is
\(
  S(P)
  = \frac{1}{|P|-1}
    \sum_{i=1}^d
      \sum_{k=1}^{|P|-1}
        \bigl(\tilde G_i(k) - \tilde G_i(k+1)\bigr)^2,
\)
where \(\tilde G_i\) is the sorted list of the \(i^{\rm th}\) objective values in \(P\), and \(\tilde G_i(k)\) is the \(k^{\rm th}\) entry in this sorted list. 
Lower sparsity indicates a more uniform distribution of solutions along each objective.

\subsection{Hungarian Matching Distance}\label{app:hungarian}

To measure structural similarity between policies, we use the Hungarian matching distance \citep{kuhn1955hungarian, munkres1957algorithms}. 
\begin{definition}
    {Hungarian matching distance.} 
    For a given layer $l$ with neuron sets $A^{(l)}$ and $B^{(l)}$, let $w_i^{(l)}$ and $w_j^{(l)}$ denote the incoming weight vectors of neurons $i \in A^{(l)}$ and $j \in B^{(l)}$, respectively. 
The minimum-cost perfect matching $M^{(l)}$ between $A^{(l)}$ and $B^{(l)}$ is obtained by using the Hungarian algorithm: 
\begin{equation}
    M^{(l)} = \underset{\text{matching}}{\arg\min} 
    \sum_{(i,j) \in \text{matching}} 
    \left\lVert w_i^{(l)} - w_j^{(l)} \right\rVert_2
\end{equation}

The Hungarian matching distance between the two networks is then defined as 
\begin{equation}
    d_{\text{Hungarian}}(A,B) = 
\sum_{l=1}^{L} 
\sum_{(i,j) \in M^{(l)}} 
\left\lVert w_i^{(l)} - w_j^{(l)} \right\rVert_2
\end{equation}
where $L$ is the total number of layers.
\end{definition}

\subsection{Sanity Check Details}\label{app:sanity_check}

In Section \ref{valhom}, the sanity check was designed as a qualitative validation of the parameter–performance relationship (PPR), rather than a full experiment.

To construct policy pairs, we trained 11 base policies $\theta_{\omega_1}$ using the scalarization vector $\omega_1$ that were equally distributed in the preference space, i.e. $\omega_1=[1,0],[0.9,0.1],...,[0,1]$. 

For each base policy, we formed a paired scalarization vector $\omega_{2}$ by shifting $\omega_{1}$ with a retraining shift parameter $\delta_{s}$:
\begin{equation}
    \omega_2=
    \begin{cases}
        [\omega_{11} - \delta_s,\omega_{12} + \delta_s], \text{ if } \omega_{11} - \delta_s \in [0,1] \\\\
        [\omega_{11} + \delta_s,\omega_{12} - \delta_s], \text{ otherwise}
    \end{cases}
\end{equation}
where $\delta_s \in \{0.1, 0.2,0.3,0.4,0.5\}$.
This procedure yielded multiple pairs of $(\omega_1,\omega_2)$ across the preference space. For each pair, we performed short retraining from $\theta_{\omega_1}$ under $\omega_2$ and compared the original $\theta_{\omega_1}$ and resulting model $\theta_{\omega'}$ with the independently trained $\theta_{\omega_2}$. 
This allowed us to consistently observe that retrained policies stay close in parameter space (low Hungarian distance), while having directional movement in performance space, supporting the local PPR hypothesis.
Figure \ref{fig:model_similarity} in the main paper presents a representative example from these trials.

\subsection{Training and Evaluation Details} 

All learning phases within our \algoname{} algorithm, including the initial training of base policies, the directional retraining, and the final preference-aligned fine-tuning, utilise the Proximal Policy Optimization (PPO) algorithm~\citep{schulman2017proximal}. We employed a standard PPO implementation from the Stable Baselines3 library~\citep{stable-baselines3}. The PPO parameters used across all training stages and benchmarks are detailed in Table~\ref{tab:ppo_params}.

The specific parameters for the \algoname{} pipeline include:
\begin{itemize}
    \item \textbf{Number of Initial Base Policies (\(K\))}: The total count of base policies \(\theta_{w_k}\), trained in the initialization stage. The corresponding \(K\) initial scalarization weights \(\{w_k\}_{k=1}^K\) are generated by evenly distributing them across the preference space (e.g., for 2D objectives, from \([1,0]\) to \([0,1]\) in \(K\) steps).
    \item \textbf{Initialisation Training Timesteps (\(T_{\mathrm{init}}\))}: The number of environment interaction steps for which the initial base policy \(\theta_{w_k}\) is trained under its weight \(w_k\).
    \item \textbf{Retraining Preference Shift Strategy (controlled by shift magnitude \(\delta_s\))}:
    Target scalarization weights \(\{w_k^{(i)}\}\) for directional retraining are generated by shifting each initial weight \(w_k\) to a nearby, distinct point on the preference space. The extent of this shift is controlled by a hyperparameter \(\delta_s\). 

    \item \textbf{Directional Retraining Timesteps (\(T_{\mathrm{dir}}\))}: The number of environment interaction steps for which the base policy \(\theta_{w_k}\) is retrained under its target weight \(w_k^{i}\) to produce \(\theta_{w_k^{(i)}}\). 
    
    \item \textbf{Step-Scale Factor Generation} (\(\alpha_{\text{start}}, \alpha_{\text{end}}, \Delta\alpha \)): The set of step-scale factors \(\{\alpha_i\}\) used in Locally Linear Extension is generated based on a starting value (\(\alpha_{\text{start}}\)), an ending value (\(\alpha_{\text{end}}\)), and either a step increment (\(\Delta\alpha\)). 
    
    \item \textbf{Fine-tuning Timesteps (\(T_{\mathrm{ref}}\))}: The number of environment interaction steps for which the selected candidate policy from the extension phase is fine-tuned under its matched preference weight \(w_{k,\text{cand}}\).
\end{itemize}

\begin{table}[h!]
    \centering
    \caption{PPO hyperparameters for benchmarks.}
    \resizebox{0.8\textwidth}{!}{%
\begin{tabular}{lccccc}
\toprule
Parameter Name           & MO-Swimmer     & MO-Hopper-2d  & MO-Ant-2d & MO-Hopper-3d  & MO-Ant-3d\\ \midrule
steps per actor batch & 512  &  512  &  512 &  512  &  512\\
learning rate (\(\times10^{-4}\))    & 3 & 3 & 3 & 3 & 3\\
learning rate decay ratio     & 1   & 1  & 1 & 1  & 1\\
$\gamma$ & 0.995   & 0.995  & 0.995 & 0.995  & 0.995\\
GAE lambda         & 0.95  & 0.95  & 0.95 & 0.95  & 0.95\\
number of mini batches  &  32  &  32 &  32 &  32 &  32\\
PPO epochs     & 10  & 10  & 10 & 10  & 10 \\
entropy coefficient    & 0.0 & 0.0 & 0.0  & 0.0 & 0.0\\
value loss coefficient & 0.5 & 0.5 & 0.5  & 0.5 & 0.5\\
maximum gradient norm          & 0.5 & 0.5 & 0.5 & 0.5 & 0.5   \\
clip parameter         & 0.2  & 0.2  & 0.2 & 0.2  & 0.2\\ \bottomrule
\end{tabular}
}
    \label{tab:ppo_params}
\end{table}

In all experiments reported in the main results, we fix the number of initial base policies to $K=6$, the retraining preference shift magnitude to $\delta_s = 0.1$, and the step-scale factor grid to $\alpha \in [-1.5, 1.5]$ with step size $\Delta\alpha = 0.05$.
We clarify that $K$ and $\delta_s$ are chosen for simplicity and are not intended
to be optimal. We have an additional discussion about the choice of $\delta_s$ in Appendix \ref{app:delta_s}. The step-scale factor $\alpha$ controls the Locally Linear Extension stage and has a meaningful but limited influence on performance; its impact and selection are discussed in Appendix \ref{app:exp_alpha}.

The number of base policies $K$ controls the initial coverage of the preference space. Too few base policies may leave some Pareto-front components unexplored, whereas increasing $K$ provides more starting points but also increases training costs. Under a fixed total interaction budget, a larger $K$ also reduces the training budget available to each base policy, so increasing $K$ does not necessarily improve performance monotonically. We use the same moderate setting $K=6$ across all two- and three-objective benchmarks, without environment-specific tuning, and consistently obtain strong Pareto-front coverage. This suggests that LLE-MORL is not highly sensitive to the precise value of $K$ within a reasonable range; additional base policies primarily improve coverage density with diminishing returns relative to their cost. 

For baseline methods, since the original papers do not evaluate performance on the same benchmark problems considered in this work, we adopt the implementations and hyperparameter settings provided by morl-baselines~\citep{felten_toolkit_2023}.
As \algoname{} is a PPO-based algorithm, we employ PPO hyperparameter configurations commonly used in PPO-based MORL methods, such as PGMORL~\citep{xu2020prediction}, which serves as a standard and widely adopted reference for PPO settings in MORL tasks.

In all experiments reported in the main results, we strictly control the total training interaction budget for all methods to ensure a fair comparison. For \algoname{}, this fixed budget is automatically partitioned across the three learning phases using a simple $3\!:\!1\!:\!1$ ratio for initial base policy training, directional retraining, and preference-aligned fine-tuning, respectively. 
When the total training budget is sufficiently large, allocating additional interaction steps to the base policy training stage can further improve performance, as directional retraining and fine-tuning require fewer updates. The optimal budget allocation depends on the target application and can be adjusted on a case-by-case basis.

Approximate execution times for all methods and benchmarks are reported separately in Table \ref{tab:runtime}. The computational resources used are described in Appendix~\ref{app:computational_resources}.

\begin{table}[t]
    \centering
    \caption{Approximate execution times (hours) for each method and benchmark under different settings.}
    \begin{subtable}{0.49\linewidth}
        \centering
        \caption{2D sample-efficient setting}
        \resizebox{\linewidth}{!}{
\begin{tabular}{l ccc}
\toprule
Method             & MO-Swimmer & MO-Hopper-2d & MO-Ant-2d  \\
\midrule
GPI-LS         &  3  &  3 &  3  \\
CAPQL         &  3  &  3  & 3 \\
Q-Pensieve &  3 &  3  &  3\\
MORL/D         &  1  &    1   &   1 \\
\algoname{}         &  1  &   1   &  1\\
\bottomrule
\end{tabular}
}

    \end{subtable}
    \begin{subtable}{0.49\linewidth}
        \centering
        \caption{2D standard-training setting}
        \resizebox{\linewidth}{!}{
\begin{tabular}{l ccc}
\toprule
Method  & MO-Swimmer & MO-Hopper-2d & MO-Ant-2d \\
\midrule
GPI-LS         & 22    &    23     &  19    \\
CAPQL         &    11    &    11  &   12   \\
Q-Pensieve &   17  & 15  & 18  \\
MORL/D         &   3   &   3   &   3    \\
\algoname{}         &   3    &   3   &  3  \\
\bottomrule
\end{tabular}
}
    \end{subtable}
    \vspace{0.1in}
    
    \begin{subtable}{0.35\linewidth}
        \centering
        \caption{3D setting}
        \resizebox{\linewidth}{!}{
\begin{tabular}{l cc}
\toprule
Method  & MO-Hopper-3d & MO-Ant-3d \\
\midrule
GPI-LS          &    44   &   43 \\
CAPQL          &  13   &   18  \\
Q-Pensieve  &  24 &  23 \\
MORL/D        &   5   &   6    \\
\algoname{}      &  5    &  5 \\
\bottomrule
\end{tabular}
}

    \end{subtable}
    
    \label{tab:runtime}
\end{table}

\paragraph{Evaluation cost.} 
Interactions with the environment for evaluation of the algorithm are usually not considered towards sample efficiency because they are not used for policy improvement. 
Nevertheless, because evaluation plays a particularly visible role in selecting LLE candidates, we understand the concern and report the evaluation cost separately here.
Under our default configuration, $\alpha \in \left[ -1.5,1.5 \right]$ at intervals of 0.05, LLE-MORL evaluates $M=61$ candidates for 5 episodes per base policy, corresponding to a total $K\times M\times\text{evaluation episodes} = 61\times6\times5 = 1830 $ evaluation episodes in the LLE stage under 2d environment. For the full algorithm, we also have base-policy evaluation, directional-retraining diagnostics, and fine-tuned policy evaluations. As the final candidate number is not fixed, the total evaluation will be approximate around 3000 episodes.
Evaluation outside the training-step budget is also standard in the compared implementations. 
For baselines: (1) CAPQL evaluates every 10,000 training steps. At each evaluation it uses $100~\text{preference weights} \times 5 ~\text{episodes} = 500 ~\text{evaluation episodes}$. Under a standard $10^6$ training budget, CAPQL has approximately 50,000 episodes of evaluation cost. (2) GPI-LS has several separate evaluation components and has at least 10,000 episodes of evaluation cost under a $10^6$ training budget. (3) MORL/D evaluates the whole population once before training and again after every 10,000 training steps. Under a $10^6$ training budget, it has 15,300 evaluation episodes. (4)~\text{Q-Pensieve} evaluates every 50,000 training steps, and it has 3,600 evaluation episodes under a $10^6$ training budget. 
These evaluation roll-outs do not update policy parameters and are separate from the common training-interaction budget.
Overall, the evaluation cost remains reasonable in the low- to moderate-dimensional settings considered here. The end-to-end wall-clock times in Table~\ref{tab:runtime} include both training and evaluation.

\subsection{Computational Resources}\label{app:computational_resources}  

All experiments were run on a workstation equipped with an AMD Ryzen Threadripper PRO 5975WX (32 cores), an NVIDIA GeForce RTX 3090 GPU (24 GB GDDR6X), and 256 GiB of RAM, running Ubuntu 24.04 LTS. The software stack included CUDA Toolkit 12.0 and the corresponding NVIDIA drivers.

\section{Additional Results}\label{app:addition_exp}

\subsection{Effect of Directional Retraining Shift}\label{app:delta_s}

This ablation study investigates the influence of the directional retraining shift \(\delta_s\), on the Locally Linear Extension process. We analyse the performance of the \algoname-0 variant (which excludes fine-tuning) under the standard-training setting. We vary \(\delta_s\) over the set \(\{0.1, 0.2, 0.3, 0.4, 0.5\}\), keeping other parameters at their default values, and report Hypervolume (HV), Expected Utility (EU), and Sparsity (SP) for each environment in Table~\ref{tab:shift_results}.

The results indicate that \(\delta_s\) influences the resulting Pareto front approximation, as different shift values generate distinct extension trajectories that directly shape the front. While optimal performance varies across environments, a moderate retraining shift (e.g., \(\delta_s\) in the range of \(0.2\) to \(0.3\), depending on the specific benchmark characteristics evident in Table~\ref{tab:shift_results}) generally appears to strike an effective balance between front coverage (HV, EU) and solution diversity (SP). Our default configuration utilised \(\delta_s=0.1\), chosen for its simplicity and promising results in preliminary tests. However, this ablation demonstrates that this typical shift is not universally optimal; selecting an appropriately tuned moderate \(\delta_s\) can further enhance the quality and coverage of the extended manifold, thereby improving the final Pareto front approximation achieved by \algoname-0.

\begin{table}[h]
    \centering
    \caption{Impact of retraining shift \(\delta_s\) on \algoname-0 performance (HV, EU, SP) under standard-training setting.}
    \resizebox{0.6\textwidth}{!}{%
\begin{tabular}{llccccc}
\toprule
\textbf{Environment} & \textbf{Metric} & \(\delta_s=0.1\) & \(\delta_s=0.2\) & \(\delta_s=0.3\) & \(\delta_s=0.4\) & \(\delta_s=0.5\) \\
\midrule
\multirow{3}{*}{MO-Swimmer}   & HV(\(10^4\)) & 7.37 & 7.38 & 7.38 & 7.36 & 7.37 \\
             & EU(\(10^1\)) & 1.03 & 1.29 & 1.41 & 0.99 & 1.60 \\
             & SP(\(10^2\)) & 2.10 & 0.83 & 3.05 & 2.39 & 2.12 \\
\midrule
\multirow{3}{*}{MO-Hopper-2d} & HV(\(10^5\)) & 4.88 & 4.84 & 4.79 & 4.86 & 4.80 \\
             & EU(\(10^2\)) & 5.65 & 5.56 & 5.49 & 5.54 & 5.58 \\
             & SP(\(10^2\)) & 7.20 & 4.49 & 3.33 & 2.54 & 3.33 \\
\midrule
\multirow{3}{*}{MO-Ant-2d}    & HV(\(10^5\)) & 2.43 & 2.66 & 2.65 & 2.72 & 2.44 \\
             & EU(\(10^2\)) & 3.44 & 3.94 & 3.69 & 3.95 & 3.66 \\
             & SP(\(10^3\)) & 1.76 & 2.64& 2.23 & 8.98 & 2.41 \\
\bottomrule
\end{tabular}%
}
    \label{tab:shift_results}
\end{table}

\subsection{Effect of Step-Scale in Locally Linear Extension}\label{app:exp_alpha}

We next examine how the choice of step-scale factors \(\{\alpha_j\}\) --- defined by their start \(\alpha_{\mathrm{start}}\), end \(\alpha_{\mathrm{end}}\), and increment \(\Delta\alpha\) --- influences the performance of \algoname-0. For each benchmark, we compare different ranges and resolutions of \(\{\alpha_j\}\) and report HV, EU, and SP in Table~\ref{tab:alpha}.

The choice of \(\Delta\alpha\) significantly impacts Sparsity (SP), as shown in Table~\ref{tab:alpha}. Finer increments (smaller \(\Delta\alpha\)) lead to substantially lower SP values, indicating denser Pareto front approximations, whereas coarser steps result in sparser solutions. This confirms that LLE offers a training-free mechanism to control the density of the approximated front simply by adjusting \(\Delta\alpha\).

Regarding the range of \(\alpha_j\) (controlled by \(\alpha_{\mathrm{start}}\) and \(\alpha_{\mathrm{end}}\)), expanding it generally leads to improvements in both HV and EU, as the extension manifold can reach further from the initial policy along the identified directional vector (visualized conceptually in Figure~\ref{fig:moving}). However, Table~\ref{tab:alpha} suggests that an extremely wide range does not always yield proportional gains in coverage. This indicates that beyond a certain point, the linearity assumption underpinning LLE may become less effective for extending the front into novel, high-quality regions using a single directional vector, or that the most valuable regions reachable by the current \(\Delta\theta\) vectors are already sufficiently captured. 

\begin{table}[t]
    \centering
    \caption{Influence of Step-Scale \(\alpha\) on \algoname-0 performance under standard-training setting.}
    \resizebox{\textwidth}{!}{%
\begin{tabular}{llc|cccc|cccc|cccc|cccc}
\toprule
\multirow{2}{*}{\textbf{Environment}}
 & \multirow{2}{*}{\textbf{Metric}} 
 & \((\alpha_{\rm start},\alpha_{\rm end})\)
 & \multicolumn{4}{c|}{(-1,1)}
 & \multicolumn{4}{c|}{(-1.5,1.5)}
 & \multicolumn{4}{c|}{(-2,2)} 
& \multicolumn{4}{c}{(-3,3)}\\
 \cmidrule(lr){3-19} 
 & & \(\Delta\alpha\)
 & 0.01 & 0.05 & 0.1 & 0.5
 & 0.01 & 0.05 & 0.1 & 0.5
& 0.01 & 0.05 & 0.1 & 0.5
 & 0.01 & 0.05 & 0.1 & 0.5\\
\midrule
\multirow{3}{*}{MO-Swimmer}   & HV(\(10^4\)) && 7.37 & 7.37 & 7.37 & 7.33 & 7.38 & 7.37 & 7.37 & 7.33 & 7.40 & 7.40 & 7.39 & 7.34 & 7.40 & 7.40 & 7.40 & 7.34 \\
             & EU(\(10^1\))&& 1.08 & 1.27 & 1.32 & 1.33 & 1.08 & 1.03 & 1.42 & 1.65 & 1.13& 1.41& 1.53& 1.70  & 1.43 & 1.75 & 1.73 & 2.37 \\
             & SP(\(10^2\))&& 0.24 & 1.17 & 2.54 & 25.55 & 1.13 & 2.10 & 1.46 & 25.55 & 0.11 & 0.53& 1.18& 11.48 & 0.11 & 0.51 & 1.18 & 11.48 \\
\midrule
\multirow{3}{*}{MO-Hopper-2d} & HV(\(10^5\)) && 4.95 & 4.88 & 4.85 & 4.80 & 4.96 & 4.88 & 4.90 & 4.80& 4.95& 4.88& 4.88& 4.81 & 4.96 & 4.88 & 4.88 & 4.81 \\
             & EU(\(10^2\)) && 5.82 & 5.74 & 5.73 & 5.75 & 5.76 & 5.65 & 5.72 & 5.75& 5.75& 5.69& 5.69& 5.67 & 5.75 & 5.69 & 5.69 & 5.68 \\
             & SP(\(10^2\)) && 5.63 & 9.43 & 11.58 & 60.35 & 5.01 & 7.20& 13.06& 60.35& 3.02& 5.42& 10.05& 29.66 & 3.62 & 5.41 & 10.05 & 29.66 \\
\midrule
\multirow{3}{*}{MO-Ant-2d}    & HV(\(10^5\))&& 2.57 & 2.43 & 2.40 & 2.23 & 2.60 & 2.43& 2.23& 2.14& 2.60& 2.47& 2.25& 2.14 & 2.60 & 2.47 & 2.25 & 2.14 \\
             & EU(\(10^2\)) && 3.61 & 3.47 & 3.62 & 3.66 & 3.66 & 3.44& 3.44& 3.31& 3.67& 3.59& 3.43& 3.37 & 3.67 & 3.59 & 3.43 & 3.37 \\
             & SP(\(10^3\)) && 0.54 & 2.03 & 3.11 & 7.92 & 1.85 & 1.76& 1.79& 8.36& 1.85& 1.55& 1.79& 8.36 & 1.85 & 1.55 & 1.79 & 8.36 \\
\bottomrule
\end{tabular}%
}
    \label{tab:alpha}
\end{table}

\section{Additional Discussion}
\subsection{Discussion of Reliability and Robustness}
In the \algoname{} algorithmic procedure, the underlying Reinforcement Learning method (PPO) used in the training and retraining phases is subject to inherent stochasticity arising from factors such as random seed initialisation for neural network weights and data sampling during training. 
This means that while the overarching behaviour and effectiveness of \algoname{} are generally reproducible, the exact set of policies discovered on the approximated Pareto front, their specific parameter values, or their ordering might exhibit some variation across independent runs started with different random seeds. Consequently, direct averaging of entire Pareto fronts or performing straightforward statistical tests on the precise composition of these policy sets can be non-trivial and may not always be the most informative way to capture the consistent ability of the algorithm to find high-quality solution regions.

Our demonstration of reliability and robustness relies on: (1) the consistent observation of the superior performance of \algoname{} in generating high-quality Pareto fronts compared to baselines; (2) sensitivity analyses of key hyperparameters (detailed in Appendix~\ref{app:addition_exp}), which show consistent outcome patterns within certain parameter ranges; (3) the qualitative consistency in the shape, extent, and dominance characteristics of the visualized Pareto fronts
(e.g.~Figure~\ref{fig:pareto_fronts_2d}); and (4) the quantitative results reported in Table \ref{tab:metrics_2d} and Table \ref{tab:metrics_3d} across multiple seeds, further reinforcing the statistical reliability of our findings.
These combined observations support the robustness of our conclusions and the effectiveness of \algoname.

\subsection{Discussion of Assumptions}

The assumption of diffeomorphic relationships or of Lipschitz conditions in the local manifold analysis provides sufficient conditions for the theoretical error bound and complexity. 
Importantly, \algoname{} only requires these properties \emph{locally}, around a base policy and a single Pareto-front branch.
In general deep RL problems, solutions are approached by gradient-like methods which assume that the loss function is locally sufficiently well approximated by a tangential space, which is a hint that diffeomorphic relationships for successfully solvable deep RL problems can be seen as natural, although  
much more work and additional (albeit less restrictive) assumptions would be needed to prove this. Also, this would not include multi-objective application cases where an optimal solution is not required nor available, but the extensions can still provide good suboptimal solutions.
Therefore, we believe that the present assumptions provide a reasonable compromise between the generality of the currently possible propositions and the coverage of practically relevant situations. 

More specifically, the analysis separates several conditions. First, Theorem~\ref{theo:lc} derives local Lipschitz regularity of the parameter--performance map from differentiability of the expected return and a bounded policy score function, conditions commonly used in RL analysis. Second, the diffeomorphism in Theorem~\ref{theo:thm:smooth_lift} concerns only a local representative parameter subset $\Theta_v$ and a neighbourhood $\mathcal{P}_v$ of one Pareto-front branch; it does not assume a global bijection mapping between all neural-network parameters and performance. Third, exact non-dominated initialisation and tangent-space recovery are theoretical conditions to express approximation error relative to the true Pareto front. 

The practical algorithm is less restrictive: the true Pareto front is unknown and the base policies are finite-budget, approximately non-dominated solutions, and brief directional retraining provides an approximation to local manifold directions rather than exact tangent vectors. Across the five continuous-control benchmarks considered, including two three-objective tasks, these local approximations consistently produced useful extension directions and high-quality Pareto-front reconstructions.

\section{Limitations\label{app_limitations}} 


Limitations of our approach are implied by inherent challenges in multi-objective optimisation, but we also note some limitations that are specific to our algorithm and require further study.

\begin{itemize}
\item 
A larger number of objectives is a problem for most of the existing MORL algorithms. Although usually some of the objectives are of different importance and can be lexicographically ranked, so that the complexity does not necessarily increase exponentially with the number of objectives. 
The high-dimensional case is nevertheless challenging, but our approach can be seen as promising: In higher dimensions, the number of solutions that are in the same local quasi-linear patch increases dramatically, so that the efficiency of the proposed local search will be even more beneficial.
This benefit could be reduced by the potentially increasing complexity of the topological relation between the performance space and the parameter space which could be a fascinating subject for future work.

\item We are assuming that the Pareto front consists of a limited number of connectivity components which have a manifold structure. 
While there is no theoretical bound to the complexity of the Pareto front, the idea of MORL implies that the objectives are at least in some sense comparable. For Pareto fronts that are fractal or of high genus, the result of multi-objective optimisation lacks robustness, although it will neither be possible to fix any limits for the complexity of the Pareto front. 
However, as long as there are only a limited number of manifold-like connectivity components, our algorithm will be applicable. 

\item Widely different scales and elasticities of the objectives can lead to problems as in optimisation in anisotropic error landscapes. Step size control that helps in gradient methods in optimisation will also be useful here, but has not been studied yet, as the typical (benchmark) problems are sufficiently isotropic. 

\item The density of the identified solutions on the Pareto front is clearly a challenge, which may be solved by step size control as mentioned in the previous point. This concerns higher-dimensional cases as well as extended one-dimensional trails as visible in the top trail in Figure~\ref{fig:moving_c} also a simple reduction of the parameter $\Delta \alpha$ at the observation of large steps in the performance space could have solved this issue already so that a more uniform covering of the Pareto from is not difficult to achieve in the present approach.  See also Appendix~\ref{app:exp_alpha}.
In contrast to other approaches, locally smooth regions of the Pareto front can trivially be tracked by \algoname. Concave regions connected to the Pareto front will be followed through without problem, but will need to be removed in a single postprocessing step as they are dominated by other solutions. Even full patches of solutions may turn out to be Pareto sub-optimal and require a similar treatment.

\item We have made use of scalarisation to seed the solution domains, whereas the reconstruction of the Pareto front is done by a lateral process that does not depend on preference weights. It is in principle possible that a solution patch is not reachable by any scalarisation-based seeding attempt, see also the early discussion in~\citep{vamplew2008limitations}. In this case our approach might not find this patch, although it is still possible that it is found by retraining from a different solution domain as shown in Figure~\ref{fig:moving_c}.   
\end{itemize}



\newpage
\section*{NeurIPS Paper Checklist}

\begin{enumerate}

\item {\bf Claims}
    \item[] Question: Do the main claims made in the abstract and introduction accurately reflect the paper's contributions and scope?
    \item[] Answer: \answerYes{} 
    \item[] Justification:  The abstract includes the contribution and the scope of developing an algorithm for efficient multi-objective reinforcement learning.

\item {\bf Limitations}
    \item[] Question: Does the paper discuss the limitations of the work performed by the authors?
    \item[] Answer: \answerYes{} 
    \item[] Justification: The limitations of the approach are discussed in Appendix~\ref{app_limitations}.

\item {\bf Theory assumptions and proofs}
    \item[] Question: For each theoretical result, does the paper provide the full set of assumptions and a complete (and correct) proof?
    \item[] Answer: \answerYes{} 
    \item[] Justification: The paper states the main theoretical assumptions and  results in Section~\ref{subsec:theoretical}, while complete proofs and supporting propositions are provided in Appendix~\ref{app:theory}.

    \item {\bf Experimental result reproducibility}
    \item[] Question: Does the paper fully disclose all the information needed to reproduce the main experimental results of the paper to the extent that it affects the main claims and/or conclusions of the paper (regardless of whether the code and data are provided or not)?
    \item[] Answer: \answerYes{} 
    \item[] Justification: The algorithm is described in detail in Section~\ref{subsec:lle_algorithm} and Algorithm~\ref{alg:pareto_extension}, and the benchmark settings, hyperparameters, training details, and computational resources are provided in Section~\ref{sec4:experiment} and Appendix~\ref{app:exp_details}.

\item {\bf Open access to data and code}
    \item[] Question: Does the paper provide open access to the data and code, with sufficient instructions to faithfully reproduce the main experimental results, as described in supplemental material?
    \item[] Answer: \answerYes{} 
    \item[] Justification: Our code is publicly available at \url{https://github.com/AQQQQY/LLE_MORL}. The simulator (MO-Gymnasium) used in the experiments has been made publicly available~\citep{felten_toolkit_2023}. 

\item {\bf Experimental setting/details}
    \item[] Question: Does the paper specify all the training and test details (e.g., data splits, hyperparameters, how they were chosen, type of optimizer) necessary to understand the results?
    \item[] Answer: \answerYes{} 
    \item[] Justification: The benchmark settings, hyperparameters, training details, and computational resources are provided in Section~\ref{sec4:experiment} and Appendix~\ref{app:exp_details}.

\item {\bf Experiment statistical significance}
    \item[] Question: Does the paper report error bars suitably and correctly defined or other appropriate information about the statistical significance of the experiments?
    \item[] Answer: \answerYes{} 
    \item[] Justification: The main result tables report mean \(\pm\) standard deviation values across runs, and Appendix~\ref{app:addition_exp} discusses robustness and sensitivity analyses of key hyperparameters.

\item {\bf Experiments compute resources}
    \item[] Question: For each experiment, does the paper provide sufficient information on the computer resources (type of compute workers, memory, time of execution) needed to reproduce the experiments?
    \item[] Answer: \answerYes{} 
    \item[] Justification: All information is given in Appendix~\ref{app:exp_details}.
    
\item {\bf Code of ethics}
    \item[] Question: Does the research conducted in the paper conform, in every respect, with the NeurIPS Code of Ethics \url{https://neurips.cc/public/EthicsGuidelines}?
    \item[] Answer: \answerYes{}{} 
    \item[] Justification: The research in this paper follows the NeurIPS Code of Ethics. The project did not work with human data or human subjects. We have considered possible misuse and have not identified any concerns for the method in its current form.

\item {\bf Broader impacts}
    \item[] Question: Does the paper discuss both potential positive societal impacts and negative societal impacts of the work performed?
    \item[] Answer: \answerNA{} 
    \item[] Justification: Without further subsequent work, we do not expect the algorithm described in the paper to be deployed in society in the near future, and thus do not discuss societal concerns at this stage.
    
\item {\bf Safeguards}
    \item[] Question: Does the paper describe safeguards that have been put in place for responsible release of data or models that have a high risk for misuse (e.g., pre-trained language models, image generators, or scraped datasets)?
    \item[] Answer: \answerNA{} 
    \item[] Justification: This paper does not pose such risks.

\item {\bf Licenses for existing assets}
    \item[] Question: Are the creators or original owners of assets (e.g., code, data, models), used in the paper, properly credited and are the license and terms of use explicitly mentioned and properly respected?
    \item[] Answer: \answerYes{} 
    \item[] Justification: The external software and benchmark environments used in this work are properly cited in the paper. MO-Gymnasium, MORL-Baselines, and Stable-Baselines3 are used in accordance with their respective open-source licenses.

\item {\bf New assets}
    \item[] Question: Are new assets introduced in the paper well documented and is the documentation provided alongside the assets?
    \item[] Answer: \answerYes{} 
    \item[] Justification: We release the implementation of LLE-MORL at \url{https://github.com/AQQQQY/LLE_MORL}, together with documentation and instructions for running the experiments.

\item {\bf Crowdsourcing and research with human subjects}
    \item[] Question: For crowdsourcing experiments and research with human subjects, does the paper include the full text of instructions given to participants and screenshots, if applicable, as well as details about compensation (if any)? 
    \item[] Answer: \answerNA{} 
    \item[] Justification: Our paper does not involve crowdsourcing nor research with human subjects.

\item {\bf Institutional review board (IRB) approvals or equivalent for research with human subjects}
    \item[] Question: Does the paper describe potential risks incurred by study participants, whether such risks were disclosed to the subjects, and whether Institutional Review Board (IRB) approvals (or an equivalent approval/review based on the requirements of your country or institution) were obtained?
    \item[] Answer: \answerNA{} 
    \item[] Justification: Our paper does not involve crowdsourcing nor research with human subjects.

\item {\bf Declaration of LLM usage}
    \item[] Question: Does the paper describe the usage of LLMs if it is an important, original, or non-standard component of the core methods in this research? Note that if the LLM is used only for writing, editing, or formatting purposes and does \emph{not} impact the core methodology, scientific rigor, or originality of the research, declaration is not required.
    \item[] Answer: \answerNA{} 
    \item[] Justification:  LLMs have been used only for writing as provided via Overleaf.

\end{enumerate}

\end{document}